\documentclass{article}

\usepackage[sglblindworkshop, final]{neurips_2026}
\workshoptitle{XAI4Science: Knowledge Discovery and Trust through Interpretable Foundation Models}

\usepackage[utf8]{inputenc}
\usepackage[T1]{fontenc}
\usepackage{hyperref}
\usepackage{url}
\usepackage{booktabs}
\usepackage{amsfonts}
\usepackage{amsmath}
\usepackage{amssymb}
\usepackage{amsthm}
\usepackage{nicefrac}
\usepackage{microtype}
\usepackage{xcolor}
\usepackage{graphicx}
\usepackage{multirow}
\usepackage{array}
\usepackage{algorithm}
\usepackage{algorithmic}

\title{
Where, What, and How to Forget: A Causal-Geometric Framework for Selective Capability Unlearning
}

\author{
  Akash Das \\
  \texttt{akash.das@fmr.com}
  \And
  Ishan Roy \\
  \texttt{ishan.roy@fmr.com}
}

\begin{document}

\maketitle

\begin{abstract}
Capability unlearning in large language models faces a fundamental tension: removing a targeted capability deeply enough to resist recovery while preserving unrelated functionality. Existing approaches largely treat unlearning as an optimization problem, modifying broad regions of parameter space without explicitly accounting for how capabilities are represented and executed. We instead formulate capability unlearning as a causal-geometric intervention problem, asking \emph{where} a capability is causally mediated, \emph{what} representational structure supports it, and \emph{how} it can be selectively modified while limiting interference. We introduce \textbf{LOCUS} (Localized Orthogonal Capability Unlearning through Subspaces), a three-stage framework combining hierarchical bidirectional causal localization, contrastive capability-subspace estimation, utility-aware geometric projection, and localized utility-orthogonal hardening. We evaluate LOCUS on Qwen3-8B, Gemma-3-4B-IT, and Phi-4-mini-instruct across biological, cybersecurity, and chemical capabilities from WMDP, alongside broad retained-utility benchmarks. Across model families, LOCUS achieves competitive forgetting while retaining substantially more general utility than global optimization baselines, with the advantage increasing under stronger forgetting. Mechanistic analyses show that capability effects are concentrated in a restricted set of components and captured by relatively low-dimensional representation structure, while cross-domain evaluations show limited interference with other hazardous capabilities. Recovery experiments further show that LOCUS consistently slows recovery under representation steering, target-domain relearning, and adversarial LoRA fine-tuning, although sufficiently strong adaptation can partially restore the capability. Overall, these results suggest that exploiting causal support and representational geometry enables more selective and recovery-resistant capability unlearning than broad parameter-space optimization.

\end{abstract}

\section{Introduction}
\label{sec:introduction}

Large language models (LLMs) can encode capabilities that are undesirable to retain or expose, including knowledge relevant to biological, chemical, and cyber misuse. While alignment and inference-time safety mechanisms regulate how such capabilities are expressed, they do not necessarily modify the underlying representations, leaving suppressed capabilities vulnerable to adversarial elicitation or subsequent adaptation. This motivates \emph{capability unlearning}: modifying a model to suppress a targeted capability while preserving unrelated functionality.

Achieving this selectively requires balancing forgetting efficacy, retained utility, and resistance to recovery. Existing optimization-based approaches can reduce target performance but often modify broad regions of parameter space, causing unnecessary utility loss, while conservative interventions may leave recoverable capability structure intact. We therefore argue that effective unlearning should account for the internal structure through which a capability is represented and executed.

We investigate three empirical hypotheses: \emph{causal sparsity}, whereby capability execution is concentrated in a restricted set of components; \emph{geometric separability}, whereby capability representations occupy relatively low-dimensional structure with limited overlap with retained utility; and \emph{local utility compatibility}, whereby forgetting can be restricted away from utility-sensitive directions. Based on these principles, we introduce \textbf{LOCUS} (\textbf{Lo}calized Orthogonal \textbf{C}apability \textbf{U}nlearning through \textbf{S}ubspaces), which combines bidirectional causal localization, contrastive capability-subspace estimation, utility-aware geometric projection, and localized hardening.

We evaluate LOCUS on WMDP-Bio, WMDP-Cyber, and WMDP-Chem across Qwen3-8B, Gemma-3-4B-IT, and Phi-4-mini-instruct, measuring target forgetting, broad retained utility, and recovery under adversarial elicitation, representation intervention, and parameter-efficient relearning. Our results show that capability effects are concentrated within restricted components and relatively low-dimensional representation structure, enabling LOCUS to achieve competitive forgetting with substantially better utility retention than global optimization baselines and consistently slower recovery.

Overall, this work reframes capability unlearning as a \emph{causal-geometric intervention problem}, showing that explicitly exploiting causal support and representational geometry can yield more selective and recovery-resistant forgetting than broad parameter-space optimization, while also revealing that sufficiently strong adaptation can still partially restore forgotten capabilities.


\section{Related Work}
\label{sec:related_work}

Machine unlearning aims to remove selected data, knowledge, or capabilities from trained models while preserving unrelated functionality. In LLMs, early approaches used post-hoc optimization to suppress targeted information \citep{eldan2023harrypotter,yao2023llmunlearning}, while TOFU \citep{maini2024tofu}, MUSE \citep{shi2025muse}, and WMDP \citep{li2024wmdp} highlight the difficulty of strong forgetting without degrading utility. Optimization-based methods modify the learning objective directly: gradient-ascent variants increase loss on forget examples, while NPO \citep{zhang2024npo} and SimNPO \citep{fan2025simnpo} target more stable forgetting--utility trade-offs. These methods primarily address \emph{how} forgetting is optimized, without identifying the internal computations that causally mediate the target capability.

A complementary line intervenes on internal representations. WMDP introduced Representation Misdirection for Unlearning (RMU), perturbing representations associated with hazardous knowledge while constraining changes on retained data \citep{li2024wmdp}. More recent methods exploit representation geometry: SSPU constructs sparse-autoencoder-derived subspaces \citep{wang2025sspu}, ReGLU combines representation geometry with retain-representation constraints \citep{xiao2026reglu}, and KUDA combines causal tracing, representation deviation, and null-space constraints \citep{fang2026kuda}. While these methods demonstrate the promise of structured representation-level intervention, an intervention subspace does not establish that its representation is causally necessary, intrinsically low-dimensional, or separable from retained capabilities. LOCUS treats these properties as empirical hypotheses rather than assumptions.

This perspective connects unlearning with mechanistic approaches to localizing model knowledge. Knowledge-neuron methods associate individual units with factual predictions \citep{dai2022knowledgeneurons}, while ROME introduced causal tracing to identify components mediating factual recall and enable localized editing \citep{meng2022rome}; MEMIT extended this paradigm to larger collections of factual associations \citep{meng2023memit}. However, attribution need not imply causal necessity: recent work on localized unlearning finds that parameters attributed to target knowledge are not necessarily uniquely required for effective forgetting \citep{lee2025localization}. LOCUS therefore separates attribution from causality, using bidirectional module-level interventions to establish causal support before constructing a sparse editable parameter set.

LOCUS also relates to geometric concept erasure and utility-preserving optimization. RLACE \citep{ravfogel2022rlace} and LEACE \citep{belrose2023leace} show that linearly represented concepts can be removed through subspace-level transformations, motivating geometric approaches to representation control. In unlearning, however, target and retained capabilities may share representational structure. LOCUS therefore estimates capability directions from matched hazardous--benign contrasts, independently estimates a utility subspace, and measures their overlap before intervening on the utility-orthogonal capability component. At the optimization level, OrthoGrad \citep{shamsian2025orthograd} and related retention-aware methods \citep{xiao2026sago} use gradient geometry to reduce interference between forgetting and retention objectives. LOCUS applies this principle locally by projecting forgetting gradients away from an estimated utility-gradient subspace within causally implicated modules.

Finally, reduced target accuracy does not necessarily imply durable removal: forgotten information can remain accessible through internal representations, activation interventions, adversarial elicitation, or subsequent fine-tuning \citep{hong2024dissecting,lee2026depth}. We therefore evaluate LOCUS along a joint forgetting--utility--recoverability frontier. The central distinction from prior work is not any individual use of localization, projection, or orthogonality, but their organization around a falsifiable causal-geometric question: \emph{do hazardous capabilities possess sufficiently sparse causal support and separable geometry to permit selective removal?} LOCUS operationalizes this through a WHERE--WHAT--HOW pipeline that causally localizes capability support, estimates its geometry relative to retained utility, and constrains intervention accordingly.


\section{Methodology: LOCUS}
\label{sec:methodology}

We formulate capability unlearning as a \emph{causal-geometric intervention
problem}. Rather than optimizing the full model against a forgetting objective,
LOCUS progressively restricts intervention to capability-relevant structure. We
test three hypotheses: \textbf{H1 (Causal Sparsity)}, that target capability
execution is disproportionately mediated by a restricted set of components;
\textbf{H2 (Geometric Separability)}, that capability representations exhibit
low-dimensional structure partially separable from retained utility; and
\textbf{H3 (First-Order Utility Preservation)}, that forgetting updates can be
restricted away from utility-sensitive parameter directions. These hypotheses
are evaluated in Sec.~\ref{sec:experiments}.

\begin{figure*}[t]
    \centering
    \includegraphics[width=\textwidth]{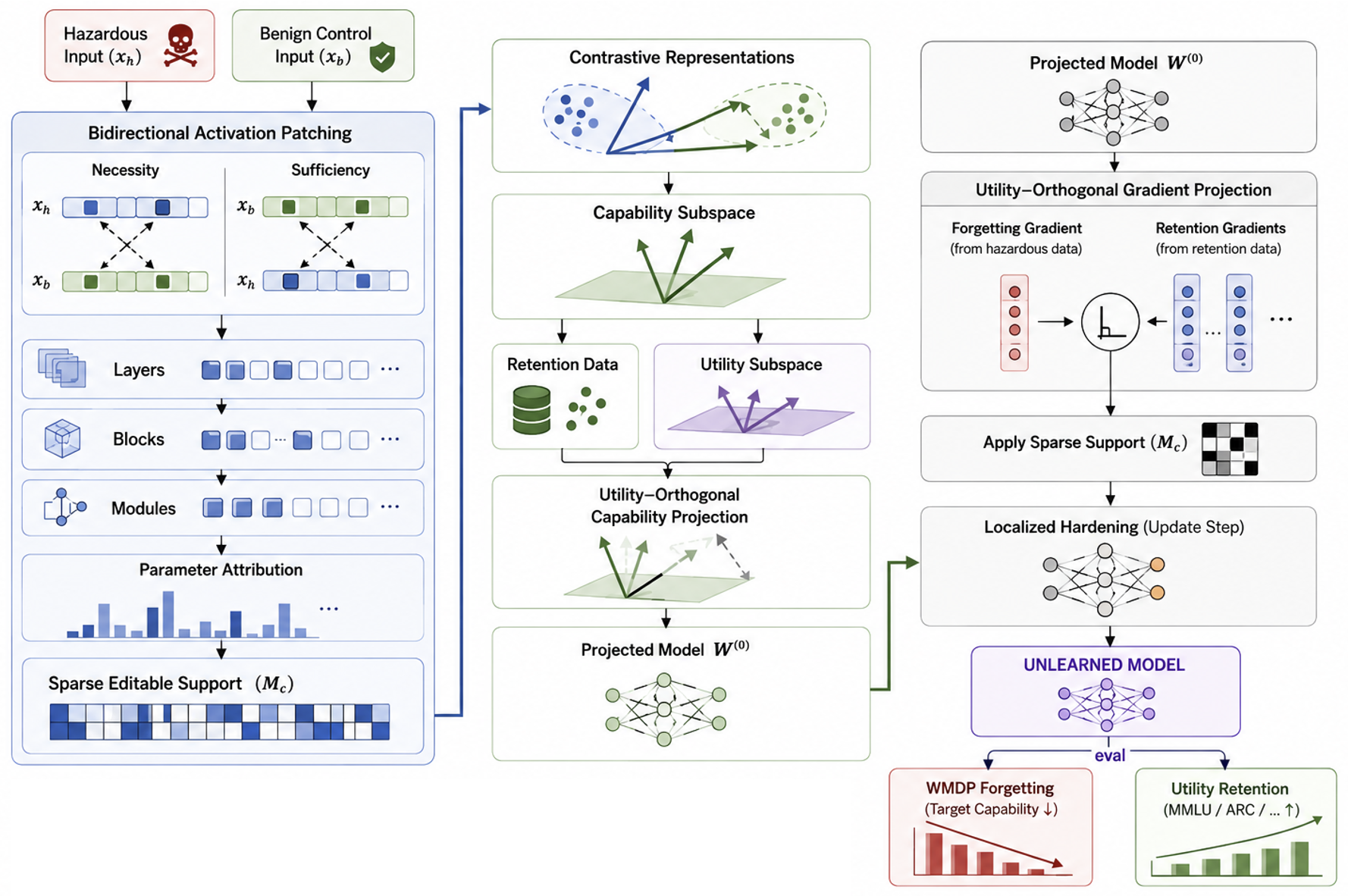}
    \caption{
    \textbf{Overview of LOCUS.}
    Given matched hazardous and benign inputs, LOCUS first identifies
    causally implicated components through bidirectional activation
    interventions and hierarchical localization. Within the resulting sparse
    support, contrastive representations define a capability subspace, while
    held-out retention data define a utility subspace. LOCUS then suppresses
    the utility-orthogonal component of the capability geometry and hardens the
    resulting model using forgetting gradients projected away from
    utility-sensitive directions and restricted to the localized support.
    }
    \label{fig:locus_overview}
\end{figure*}

As shown in Fig.~\ref{fig:locus_overview}, LOCUS proceeds through
\emph{causal localization}, \emph{capability geometry}, and
\emph{utility-constrained hardening}. Given a hazardous input $x_h$, we
construct a matched benign control $x_b$ that preserves topic and context
while changing the target intent. Let $h_l^h$ and $h_l^b$ denote their
activations at candidate module $l$, and let $s_h(x)$ denote the capability
score, defined as the correct-answer logit margin for WMDP. We measure
necessity and sufficiency through bidirectional activation interventions:

\begin{equation}
\label{eq:causal_score}
C_l =
\underbrace{
\mathbb{E}
\left[
s_h(x_h)
-
s_h\left(x_h; h_l^h \leftarrow h_l^b\right)
\right]
}_{C_l^{\mathrm{nec}}}
+
\beta
\underbrace{
\mathbb{E}
\left[
s_h\left(x_b; h_l^b \leftarrow h_l^h\right)
-
s_h(x_b)
\right]
}_{C_l^{\mathrm{suf}}}.
\end{equation}

Here, $\beta$ weights the sufficiency term. LOCUS applies these interventions
hierarchically, from layers to attention/MLP blocks and individual projection
modules, retaining implicated components at each level. Parameter-level
attribution within the selected modules defines the sparse editable mask
$M_{\mathcal C}$, which specifies the support for subsequent updates.

Within this support, we estimate capability representation geometry. For a
localized linear module $z_l = W_l h_{l-1}$, let $a_h^{(i)}$ and $a_b^{(i)}$
denote the inputs for the $i$th matched pair. Their contrastive displacement is

\begin{equation}
\label{eq:contrastive_displacement}
\delta_i =
a_h^{(i)} - a_b^{(i)}.
\end{equation}

Across $N$ matched pairs, we estimate the contrastive covariance and its
eigendecomposition:

\begin{equation}
\label{eq:contrastive_covariance}
C_{\Delta,l}
=
\frac{1}{N}
\sum_{i=1}^{N}
\delta_i \delta_i^{\top},
\qquad
C_{\Delta,l}
=
V_l \Lambda_l V_l^{\top}.
\end{equation}

The leading $r$ eigenvectors define the capability basis
$V_{h,l} = [v_1,\ldots,v_r]$. We vary $r$ experimentally. An independent
utility basis $V_{u,l}$ is estimated from held-out retention data at the same
module. Their geometric overlap is

\begin{equation}
\label{eq:geometric_overlap}
\Omega(H,U)
=
\frac{1}{r_h}
\left\|
V_{u,l}^{\top} V_{h,l}
\right\|_F^2
=
\frac{1}{r_h}
\sum_{i=1}^{r_h}
\cos^2 \theta_i.
\end{equation}

Here, $\theta_i$ are the principal angles between the capability and utility
subspaces. LOCUS isolates the utility-orthogonal component of the capability
geometry:

\begin{equation}
\label{eq:representation_intervention}
Q_{h,l}
=
\operatorname{orth}
\left[
\left(
I -
V_{u,l} V_{u,l}^{\top}
\right)
V_{h,l}
\right],
\qquad
W_l^{(0)}
=
W_l
\left(
I -
\lambda Q_{h,l} Q_{h,l}^{\top}
\right).
\end{equation}

Here, $\lambda \in [0,1]$ controls intervention strength. Since the geometry
is defined in the input space of $W_l$, the projection acts by right
multiplication.

Finally, LOCUS performs localized hardening to suppress residual or alternative
pathways. Utility gradients from independent retention minibatches define a
low-rank orthonormal basis $Q_{u,l}$ of utility-sensitive parameter
directions. The forgetting gradient is

\begin{equation}
\label{eq:forgetting_gradient}
g_{h,l}
=
\nabla_{W_l}
\mathcal{L}_{\mathrm{forget}}.
\end{equation}

We project the forgetting gradient away from the utility-gradient span:

\begin{equation}
\label{eq:orthogonal_gradient}
\operatorname{vec}
\left(
g_{h,l}^{\perp}
\right)
=
\left(
I -
Q_{u,l} Q_{u,l}^{\top}
\right)
\operatorname{vec}
\left(
g_{h,l}
\right).
\end{equation}

The resulting update is restricted to the localized support:

\begin{equation}
\label{eq:locus_update}
W_l^{(t+1)}
=
W_l^{(t)}
-
\eta
\left(
M_{\mathcal C,l}
\odot
g_{h,l}^{\perp}
\right).
\end{equation}

Here, $\eta$ is the hardening learning rate. The mask $M_{\mathcal C,l}$
determines \emph{where} parameters may change, while gradient projection
determines \emph{which directions} are removed. The first-order motivation is

\begin{equation}
\label{eq:first_order_utility}
\mathcal{L}_u(W + \Delta W)
\approx
\mathcal{L}_u(W)
+
\nabla_W \mathcal{L}_u^{\top}
\Delta W.
\end{equation}

Hence, an update orthogonal to the estimated utility-gradient span suppresses
the corresponding first-order utility change. This is not a guarantee because
estimation error and higher-order effects can introduce residual interference,
which we evaluate in Sec.~\ref{sec:experiments}.

Overall, LOCUS progressively narrows intervention from \emph{causal support},
to \emph{capability-specific geometry}, to \emph{utility-compatible updates}.
Complete derivations, implementation details, and the masking--projection
interaction are provided in Appendix~\ref{app:methodology}.

\section{Experimental Setup}
\label{sec:experiments}

We evaluate LOCUS on three independently developed small language-model
families: \textbf{Qwen3-8B}, \textbf{Gemma-3-4B-IT}, and
\textbf{Phi-4-mini-instruct}, with Qwen3-8B serving as the primary model for
causal and geometric analyses and the other models providing
cross-family replication. We use the \textbf{Weapons of Mass Destruction
Proxy (WMDP)} benchmark~\citep{li2024wmdp} as the primary hazardous-capability
evaluation, treating WMDP-Bio, WMDP-Cyber, and WMDP-Chem as separate
unlearning targets. Matched benign controls preserve topic and prompt
structure while replacing the hazardous intent, enabling causal intervention
and contrastive representation analysis. Retained utility is evaluated using
MMLU, ARC-Challenge, HellaSwag, TruthfulQA, and WikiText-2, with results
reported relative to each model's original performance. All intervention and
evaluation data are partitioned to prevent reuse of evaluation examples
during localization, geometry estimation, or hardening. Experiments examine
four complementary properties of LOCUS: target-capability forgetting,
retained utility, causal concentration and geometric structure, and
post-unlearning recovery under adversarial intervention or adaptation. All
models retain their original architectures and parameters except for the
localized LOCUS updates. Detailed dataset partitions, matched-control
construction, model configurations, batch size, precision, optimization
settings, hyperparameters, and evaluation protocols are provided in
Appendix~\ref{app:experimental_details}.
\section{Results and Discussion}
\label{sec:results}

We evaluate our approach along four complementary dimensions: the
forgetting--utility trade-off, causal concentration of capability-relevant
computation, geometric structure enabling selective intervention, and
resistance to post-unlearning recovery. Together, these experiments assess
both the empirical effectiveness of LOCUS and the structural hypotheses
underlying its causal-geometric formulation.

\subsection{Selective Capability Removal and the Forgetting--Utility Trade-off}
\label{sec:results_main}

We first evaluate whether LOCUS suppresses the targeted hazardous capability
while preserving unrelated functionality. Table~\ref{tab:main_results}
compares LOCUS with global gradient ascent (GA), gradient difference (GD), and
representation-level intervention across the three model families. Lower WMDP
accuracy indicates stronger forgetting, while higher utility retention is
better. Values are reported as mean $\pm$ standard deviation across
independent runs.

\begin{table*}[t]
\centering
\small
\setlength{\tabcolsep}{4.8pt}
\caption{
\textbf{Forgetting--utility trade-off across model families.}
Lower WMDP accuracy indicates stronger target-capability forgetting
($\downarrow$), while higher retained utility is better ($\uparrow$).
Avg.\ WMDP denotes the average across Bio, Cyber, and Chem. Utility retention
is normalized relative to each original model. Values report mean $\pm$
standard deviation across independent runs.
}
\label{tab:main_results}
\begin{tabular}{llccccc}
\toprule
\textbf{Model} & \textbf{Method}
& \textbf{Bio $\downarrow$}
& \textbf{Cyber $\downarrow$}
& \textbf{Chem $\downarrow$}
& \textbf{Avg. WMDP $\downarrow$}
& \textbf{Utility Ret. $\uparrow$} \\
\midrule

Qwen3-8B
& Original
& $64.8 \pm 0.4$
& $61.7 \pm 0.5$
& $59.9 \pm 0.4$
& $62.1 \pm 0.3$
& $100.0$ \\

& Global GA
& $38.9 \pm 1.2$
& $40.8 \pm 1.0$
& $39.6 \pm 1.3$
& $39.8 \pm 0.9$
& $87.4 \pm 1.5$ \\

& Gradient Difference
& $40.7 \pm 1.0$
& $42.1 \pm 1.2$
& $41.0 \pm 0.9$
& $41.3 \pm 0.8$
& $91.2 \pm 1.1$ \\

& Representation-based
& $42.3 \pm 1.1$
& $43.6 \pm 0.9$
& $42.7 \pm 1.3$
& $42.9 \pm 0.8$
& $94.0 \pm 0.9$ \\

& \textbf{Ours}
& $\mathbf{37.8 \pm 0.9}$
& $\mathbf{39.9 \pm 1.1}$
& $\mathbf{38.9 \pm 1.0}$
& $\mathbf{38.9 \pm 0.7}$
& $\mathbf{95.8 \pm 0.8}$ \\
\midrule

Gemma-3-4B-IT
& Original
& $60.9 \pm 0.5$
& $58.1 \pm 0.4$
& $57.2 \pm 0.6$
& $58.7 \pm 0.3$
& $100.0$ \\

& Global GA
& $\mathbf{38.6 \pm 1.4}$
& $41.4 \pm 1.2$
& $40.5 \pm 1.1$
& $40.2 \pm 0.9$
& $86.8 \pm 1.7$ \\

& Gradient Difference
& $41.2 \pm 1.1$
& $42.6 \pm 1.0$
& $41.9 \pm 1.3$
& $41.9 \pm 0.8$
& $90.5 \pm 1.3$ \\

& Representation-based
& $42.8 \pm 1.0$
& $44.0 \pm 1.2$
& $43.1 \pm 1.1$
& $43.3 \pm 0.8$
& $93.2 \pm 1.0$ \\

& \textbf{Ours}
& $\mathbf{39.0 \pm 1.0}$
& $\mathbf{40.6 \pm 0.9}$
& $\mathbf{40.0 \pm 1.2}$
& $\mathbf{39.9 \pm 0.7}$
& $\mathbf{95.1 \pm 0.9}$ \\
\midrule

Phi-4-mini-instruct
& Original
& $58.6 \pm 0.6$
& $56.3 \pm 0.5$
& $55.1 \pm 0.5$
& $56.7 \pm 0.4$
& $100.0$ \\

& Global GA
& $40.5 \pm 1.3$
& $42.2 \pm 1.4$
& $41.3 \pm 1.1$
& $41.3 \pm 0.9$
& $88.1 \pm 1.4$ \\

& Gradient Difference
& $41.9 \pm 1.0$
& $\mathbf{40.9 \pm 1.3}$
& $42.5 \pm 1.2$
& $41.8 \pm 0.9$
& $91.4 \pm 1.2$ \\

& Representation-based
& $43.1 \pm 1.2$
& $44.2 \pm 1.0$
& $43.7 \pm 1.4$
& $43.7 \pm 0.9$
& $93.7 \pm 1.0$ \\

& \textbf{Ours}
& $\mathbf{39.8 \pm 1.1}$
& $\mathbf{41.4 \pm 1.0}$
& $\mathbf{40.8 \pm 0.9}$
& $\mathbf{40.7 \pm 0.7}$
& $\mathbf{95.5 \pm 0.8}$ \\

\bottomrule
\end{tabular}
\end{table*}

Across all three model families, LOCUS achieves a favorable
forgetting--utility trade-off. On Qwen3-8B, average WMDP accuracy decreases
from $62.1\%$ to $38.9\%$ while retaining $95.8\%$ of aggregate utility,
compared with $87.4\%$ for Global GA at a similar forgetting level. The same
pattern holds for Gemma-3-4B-IT and Phi-4-mini-instruct, where LOCUS retains
$95.1\%$ and $95.5\%$ of baseline utility, respectively, while maintaining
competitive forgetting across all three hazardous domains.

The primary advantage is therefore not simply stronger forgetting, but
competitive target suppression with substantially less collateral degradation.
This cross-family consistency motivates the causal and geometric analyses
that follow. Full benchmark-level utility results are reported in
Appendix~\ref{app:additional_results}.

\subsection{Target Capabilities Exhibit Concentrated Causal Support}
\label{sec:results_causal}

\begin{figure*}[t]
    \centering
    \includegraphics[width=0.88\textwidth]{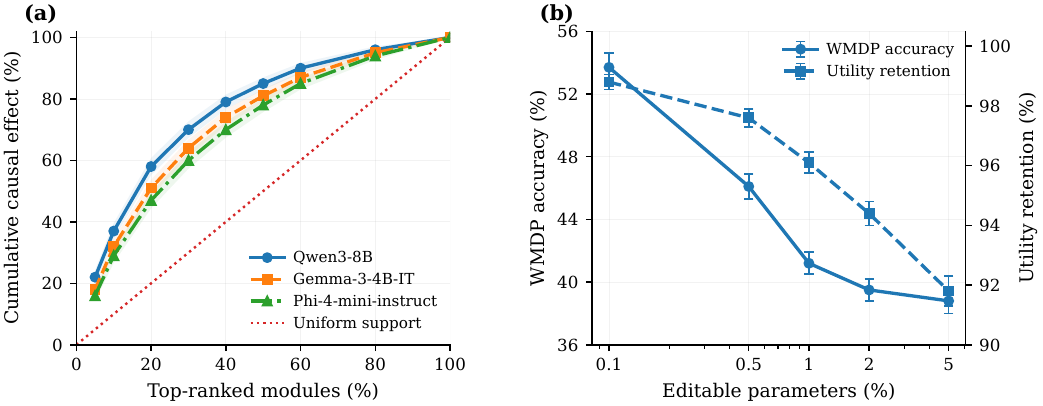}
    \caption{
    \textbf{Causal concentration and sufficiency of localized support.}
    \textbf{(a)} Cumulative causal effect captured by increasingly large
    fractions of top-ranked modules, with module-level causal scores aggregated
    across the three WMDP domains within each model family. The diagonal denotes
    the expected profile under uniformly distributed causal support.
    \textbf{(b)} Forgetting--utility behavior as the causally selected editable
    parameter fraction is increased for Qwen3-8B. Error bars denote standard deviation across the three main stochastic runs.
    }
    \label{fig:causal_sparsity}
\end{figure*}

We next test H1 by asking whether capability-relevant computation is
disproportionately mediated by a restricted set of model components.
Figure~\ref{fig:causal_sparsity} examines both the concentration of causal
influence and the editable support required for effective forgetting.

Figure~\ref{fig:causal_sparsity}(a) shows a strongly non-uniform distribution
of causal influence when module-level causal scores are aggregated across the
three WMDP domains within each model. The top 20\% of ranked modules account for
approximately 58\%, 51\%, and 47\% of cumulative causal effect in Qwen3-8B,
Gemma-3-4B-IT, and Phi-4-mini-instruct, respectively, while the top 40\%
account for approximately 79\%, 74\%, and 70\%. The convergence of the curves
toward 100\% nevertheless indicates a distributed tail rather than a strictly
localized circuit.

This concentration translates into useful sparse intervention. As shown in
Figure~\ref{fig:causal_sparsity}(b), increasing the editable fraction from
0.1\% to 1\% reduces average WMDP accuracy on Qwen3-8B from 53.7\% to 41.2\%
while retaining 96.1\% utility. Expanding the support to 5\% produces only a
further 2.4-point reduction, to 38.8\%, while utility falls to 91.8\%.
Thus, most attainable forgetting is obtained from a relatively small causal
support, with diminishing returns from broader modification.

These results support a qualified form of H1: capability execution is not
strictly localized, but its causal influence is sufficiently concentrated to
enable substantially sparser intervention. Cross-model support and
hyperparameter sweeps are reported in Appendix~\ref{app:sensitivity}, with
additional domain-level localization results in
Appendix~\ref{app:additional_results}.

\subsection{Capability Structure Is Low-Dimensional and Partially Separable from Utility}
\label{sec:results_geometry}

\begin{figure*}[t]
    \centering
    \includegraphics[width=\textwidth]{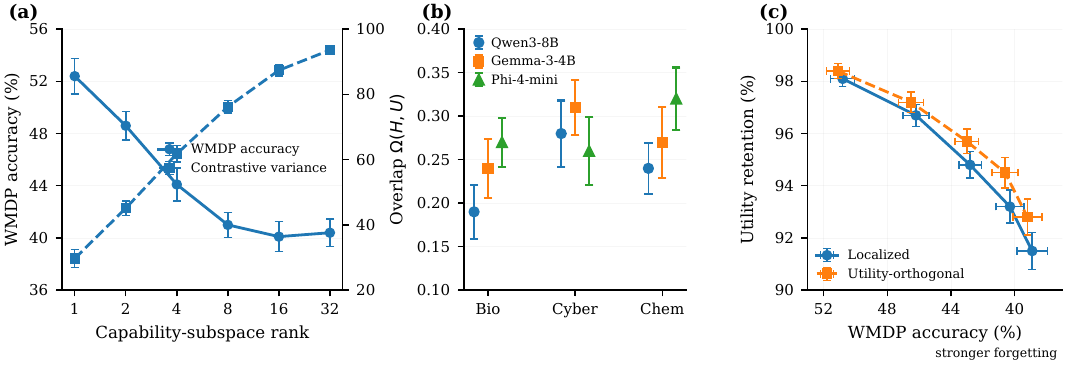}
    \caption{
    \textbf{Geometric structure and utility-aware intervention.}
    \textbf{(a)} WMDP accuracy and cumulative contrastive variance as the
    estimated capability-subspace rank is increased for the representative
    model-level geometric analysis. \textbf{(b)} Capability--utility
    subspace overlap across hazardous domains and model families.
    \textbf{(c)} Forgetting--utility trade-off under increasing intervention
    strength. Utility-orthogonal updates generally retain more utility than
    localized updates at comparable WMDP accuracy. Error bars denote standard deviation across the three main stochastic runs.
    }
    \label{fig:capability_geometry}
\end{figure*}

We next test whether capability representations exhibit the compact and
partially separable geometry required for selective intervention.
Figure~\ref{fig:capability_geometry} summarizes the rank,
capability--utility overlap, and utility-orthogonalization analyses.

Figure~\ref{fig:capability_geometry}(a) shows that the capability
representation is effectively low-dimensional. Increasing the subspace rank
from 1 to 8 reduces WMDP accuracy from approximately $52.4\%$ to $41.0\%$,
while the corresponding directions explain roughly $76\%$ of contrastive
variance. Increasing the rank to 16 reduces accuracy only to $40.1\%$, with
marginal change at rank 32, indicating that most useful intervention is
captured by a compact set of directions.

Figure~\ref{fig:capability_geometry}(b) shows that capability and utility
representations are partially, rather than completely, separated. The
normalized overlap $\Omega(H,U)$ ranges from approximately $0.19$ to $0.32$
across models and domains. This partial separation provides room for selective
intervention while explaining why stronger forgetting can still affect
retained functionality.

Finally, Figure~\ref{fig:capability_geometry}(c) shows the benefit of
utility-aware orthogonalization. Around $43\%$ WMDP accuracy, utility
retention is approximately $94.8\%$ for the localized update versus $95.7\%$
after utility orthogonalization, with the difference increasing under stronger
intervention. Together, these results support the geometric assumptions
underlying LOCUS: capability structure is compact but partially entangled
with retained functionality, making explicit control of utility-sensitive
directions beneficial. Cross-model rank and intervention-strength analyses are
reported in Appendix~\ref{app:sensitivity}.

\subsection{Forgetting Remains More Resistant to Post-Unlearning Recovery}
\label{sec:results_recovery}

Finally, we test whether comparable initial forgetting leads to different
recovery behavior. We consider representation steering and short target-domain
relearning, beginning from approximately matched $40\%$ WMDP accuracy.

\begin{figure}[t]
    \centering
    \includegraphics[width=\linewidth]{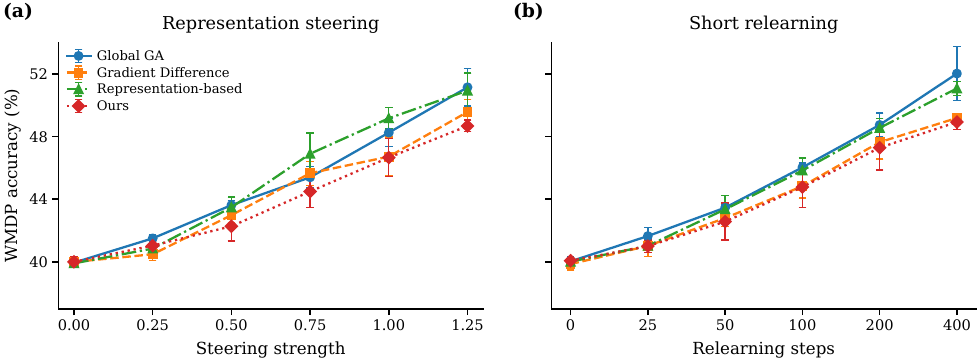}
    \caption{
    \textbf{Resistance to post-unlearning capability recovery.}
    All methods are evaluated from approximately matched initial WMDP accuracy.
    \textbf{(a)} Recovery under increasing representation-steering strength.
    \textbf{(b)} Recovery during short target-domain relearning.
    Higher WMDP accuracy indicates greater recovery of the hazardous capability.
    All approaches exhibit recovery under sufficiently strong intervention,
    while LOCUS generally recovers more slowly at stronger intervention levels.
    Error bars denote standard deviation across the three main stochastic runs.
    }
    \label{fig:recovery}
\end{figure}

Figure~\ref{fig:recovery}(a) shows that all methods progressively recover
target-domain performance as representation-steering strength increases.
Differences are small under weak steering, but LOCUS exhibits slower recovery
than the representation-based and global optimization baselines at stronger
steering levels.

A similar pattern appears under short target-domain relearning in
Figure~\ref{fig:recovery}(b). Recovery occurs for all methods, but LOCUS
remains below the strongest-recovering baselines at later checkpoints. Its
WMDP accuracy nevertheless increases substantially with additional
relearning, indicating delayed rather than prevented recovery.

Thus, comparable initial forgetting can conceal meaningful differences in
recoverability. LOCUS increases resistance to the tested recovery mechanisms
by restricting intervention to causally localized support and accounting for
utility-sensitive geometry. However, because the capability remains
recoverable under sufficiently strong intervention, these results support
\emph{increased resistance to recovery} rather than irreversible capability
deletion. A separate adversarial LoRA fine-tuning stress test is reported in
Appendix~\ref{app:relearning_recovery}.


\section{Conclusion}
\label{sec:conclusion}

We studied capability unlearning through a causal-geometric lens, asking where hazardous capabilities are mediated, what structure supports them, and how they can be modified without unnecessary collateral damage. This led to LOCUS, combining causal localization, capability-subspace estimation, and utility-orthogonal hardening for selective unlearning. Across three model families and hazardous domains, LOCUS retains approximately 95\% of measured general utility while reducing wall-clock cost by approximately $34\times$ over full-parameter retraining. Our analyses show that capabilities are concentrated in a minority of components and relatively low-dimensional subspaces, with limited overlap with retained utility, while recovery remains slower under tested adaptation mechanisms. These findings suggest that understanding capability structure can enable more selective and recovery-resistant unlearning, motivating causal and geometric analysis for targeted interventions in capable language models.
These results nevertheless leave important questions open. Our evaluation is
centered on WMDP and a limited set of retained-utility benchmarks, and we do not
experimentally compare against every recent selective-unlearning method,
particularly approaches introduced after our experimental design. The
localization and geometric analyses also introduce additional computational
cost, while recovery through arbitrary future adaptation mechanisms cannot be
ruled out. Broader capability domains and permanent deletion therefore remain
open directions. Overall, our results suggest that causal localization and
representational geometry can provide useful guides for selective and
recovery-resistant capability unlearning.

%
% ...

\newpage
\bibliographystyle{plainnat}
\bibliography{references}

\newpage
\section*{NeurIPS Paper Checklist}

\begin{enumerate}

\item {\bf Claims}
    \item[] Question: Do the main claims made in the abstract and introduction accurately reflect the paper's contributions and scope?
    \item[] Answer: \answerYes{}
    \item[] Justification: The abstract and Section 1 state the paper's main contributions, research hypotheses, and scope. The three hypotheses concerning causal sparsity, geometric separability, and first-order utility preservation are explicitly framed as empirical questions rather than universal assumptions. Their evaluation is reported in Sections 4--5 and supported by the detailed analyses in Appendix C.

\item {\bf Limitations}
    \item[] Question: Does the paper discuss the limitations of the work performed by the authors?
    \item[] Answer: \answerYes{}
    \item[] Justification: The paper discusses limitations concerning the evaluated capability domains and model scale, computational overhead, the distinction between capability suppression and permanent deletion, and the limits of the first-order utility-preservation argument. These limitations are discussed in Section 6 and further formalized through failure conditions and recovery analyses in Appendix A and Appendix C.

\item {\bf Theory assumptions and proofs}
    \item[] Question: For each theoretical result, does the paper provide the full set of assumptions and a complete (and correct) proof?
    \item[] Answer: \answerYes{}
    \item[] Justification: The mathematical assumptions, definitions, derivations, and theoretical arguments underlying the capability and utility subspaces, geometric projection, utility-gradient projection, and first-order utility preservation are provided in Section 3 and the corresponding derivations in Appendix A. The paper also explicitly discusses the limitations introduced by estimation error, higher-order effects, and the interaction between masking and projection.

\item {\bf Experimental result reproducibility}
    \item[] Question: Does the paper fully disclose all the information needed to reproduce the main experimental results of the paper to the extent that it affects the main claims and/or conclusions of the paper (regardless of whether the code and data are provided or not)?
    \item[] Answer: \answerYes{}
    \item[] Justification: The paper provides the experimental protocol, model configurations, dataset construction and partitioning, matched-control construction, evaluation procedures, optimization settings, and hyperparameters in Section 4 and Appendix B. Algorithm 1 and the methodological specifications in Appendix A provide the additional procedural detail required to reproduce the LOCUS pipeline.

\item {\bf Open access to data and code}
    \item[] Question: Does the paper provide open access to the data and code, with sufficient instructions to faithfully reproduce the main experimental results, as described in supplemental material?
    \item[] Answer: \answerNo{}
    \item[] Justification: Source code, evaluation scripts, and model checkpoints are not included with the submission. However, the paper provides detailed algorithmic specifications, mathematical formulations, experimental configurations, and descriptions of the public benchmark assets used. The source code and associated evaluation materials will be publicly released upon acceptance.

\item {\bf Experimental setting/details}
    \item[] Question: Does the paper specify all the training and test details (e.g., data splits, hyperparameters, how they were chosen, type of optimizer) necessary to understand the results?
    \item[] Answer: \answerYes{}
    \item[] Justification: Section 4 and Appendix B specify the evaluated model families, hazardous-capability datasets, matched-control construction, data separation, retained-utility evaluation, optimization procedure, hyperparameters, rank-selection procedures, recovery protocols, and evaluation settings. The appendices additionally document the implementation and computational configurations used for the experiments.

\item {\bf Experiment statistical significance}
    \item[] Question: Does the paper report error bars suitably and correctly defined or other appropriate information about the statistical significance of the experiments?
    \item[] Answer: \answerYes{}
    \item[] Justification: Quantitative results are reported as means with standard deviations across three independent random seeds. Error bars are provided for the corresponding experimental results, with numerical results summarized in the main and appendix tables and accompanying figures.

\item {\bf Experiments compute resources}
    \item[] Question: For each experiment, does the paper provide sufficient information on the computer resources (type of compute workers, memory, time of execution) needed to reproduce the experiments?
    \item[] Answer: \answerYes{}
    \item[] Justification: Appendix B documents the implementation and hardware configuration, including GPU type, precision, and execution settings, while Appendix C reports phase-wise wall-clock runtime and peak GPU memory consumption across the evaluated model families. The computational benchmark also compares LOCUS with the principal baseline methods and full-parameter retraining.

\item {\bf Code of ethics}
    \item[] Question: Does the research conducted in the paper conform, in every respect, with the NeurIPS Code of Ethics \url{https://neurips.cc/public/EthicsGuidelines}?
    \item[] Answer: \answerYes{}
    \item[] Justification: The authors have reviewed the NeurIPS Code of Ethics and conducted the study in accordance with its principles, including responsible use of public models and benchmark data, appropriate evaluation practices, and consideration of the potential dual-use implications of capability unlearning.

\item {\bf Broader impacts}
    \item[] Question: Does the paper discuss both potential positive societal impacts and negative societal impacts of the work performed?
    \item[] Answer: \answerYes{}
    \item[] Justification: Section 1 and Section 6 discuss the potential benefits of selective capability unlearning for open-weight model safety and dual-use risk mitigation, as well as potential negative implications associated with the use of unlearning techniques. The discussion also distinguishes capability suppression from permanent deletion and considers the possibility of adversarial recovery.

\item {\bf Safeguards}
    \item[] Question: Does the paper describe safeguards that have been put in place for responsible release of data or models that have a high risk for misuse (e.g., pre-trained language models, image generators, or scraped datasets)?
    \item[] Answer: \answerNA{}
    \item[] Justification: The study does not introduce or release a new foundation model or high-risk dataset. Experiments are conducted using existing publicly released open-weight language models and established benchmark datasets.

\item {\bf Licenses for existing assets}
    \item[] Question: Are the creators or original owners of assets (e.g., code, data, models), used in the paper, properly credited and are the license and terms of use explicitly mentioned and properly respected?
    \item[] Answer: \answerYes{}
    \item[] Justification: The paper identifies and cites the publicly released language models and benchmark datasets used in the experiments, including Qwen3-8B, Gemma-3-4B-IT, Phi-4-mini-instruct, and the WMDP and retained-utility benchmarks. Their use follows the associated public licensing and terms of use described or referenced in Section 4 and Appendix B.

\item {\bf New assets}
    \item[] Question: Are new assets introduced in the paper well documented and is the documentation provided alongside the assets?
    \item[] Answer: \answerNA{}
    \item[] Justification: The paper does not introduce a new standalone dataset, benchmark, or foundation model. The proposed LOCUS procedure is a methodological contribution evaluated using existing public models and benchmark assets.

\item {\bf Crowdsourcing and research with human subjects}
    \item[] Question: For crowdsourcing experiments and research with human subjects, does the paper include the full text of instructions given to participants and screenshots, if applicable, as well as details about compensation (if any)?
    \item[] Answer: \answerNA{}
    \item[] Justification: The study does not involve crowdsourcing, human participants, or experiments requiring participant instructions or compensation.

\item {\bf Institutional review board (IRB) approvals or equivalent for research with human subjects}
    \item[] Question: Does the paper describe potential risks incurred by study participants, whether such risks were disclosed to the subjects, and whether Institutional Review Board (IRB) approvals (or an equivalent approval/review based on the requirements of your country or institution) were obtained?
    \item[] Answer: \answerNA{}
    \item[] Justification: The study does not involve human participants, personal data collection, or human-subject experiments requiring IRB or equivalent institutional approval.

\item {\bf Declaration of LLM usage}
    \item[] Question: Does the paper describe the usage of LLMs if it is an important, original, or non-standard component of the core methods in this research? Note that if the LLM is used only for writing, editing, or formatting purposes and does not impact the core methodology, scientific rigor, or originality of the research, declaration is not required.
    \item[] Answer: \answerNA{}
    \item[] Justification: Language models are the objects of empirical study in this work, but their use as evaluated model subjects is not itself an original or non-standard component of the LOCUS algorithm. The core methodology consists of causal localization, contrastive representation geometry, and localized utility-orthogonal hardening as described in Section 3.

\end{enumerate}

\newpage
\appendix
\section*{Appendix}

\section{Detailed LOCUS Methodology}
\label{app:methodology}

This appendix provides the complete mathematical specification and
implementation details of LOCUS. Following the organization of the main
methodology, LOCUS determines \emph{where} the target capability is causally
mediated, \emph{what} representational structure distinguishes it from
retained functionality, and \emph{how} the localized model is modified while
limiting utility interference.

\subsection{Problem Formulation}
\label{app:problem_formulation}

Let a pretrained language model be parameterized by
\[
W=\{W_1,\ldots,W_L\},
\]
and let $\mathcal{D}_h$ denote examples expressing a target hazardous
capability and $\mathcal{D}_u$ denote retention data representing
functionality that should remain unchanged. We seek an intervened model
$W'=W+\Delta W$ that suppresses the target capability while preserving
performance on $\mathcal{D}_u$.

A generic capability-unlearning objective is
\begin{equation}
\label{eq:app_generic_unlearning}
\max_{\Delta W}
\mathcal{L}_{\mathrm{forget}}(W+\Delta W)
\qquad
\text{s.t.}
\qquad
\mathcal{L}_{u}(W+\Delta W)
-
\mathcal{L}_{u}(W)
\leq\epsilon.
\end{equation}

LOCUS does not optimize this objective globally. Instead, it tests whether the
feasible intervention can be restricted using causal and geometric structure
of the target capability.

\paragraph{H1: Causal sparsity.}
There exists a localized support $\mathcal{C}_h$ that mediates a
disproportionate fraction of target-capability execution:
\begin{equation}
\label{eq:app_h1}
\frac{|\mathcal{C}_h|}{|W|}\ll1.
\end{equation}

\paragraph{H2: Geometric separability.}
Within causally implicated modules, capability representations exhibit
comparatively low intrinsic dimensionality,
\begin{equation}
\label{eq:app_h2_rank}
\operatorname{rank}(\mathcal{H}_h)\ll d,
\end{equation}
and limited overlap with retained utility,
\begin{equation}
\label{eq:app_h2_overlap}
\Omega(\mathcal{H}_h,\mathcal{U})\ll1.
\end{equation}

\paragraph{H3: First-order utility preservation.}
Forgetting updates projected away from utility-sensitive gradient directions
should have a reduced first-order effect on the retention objective:
\begin{equation}
\label{eq:app_h3}
\nabla_W\mathcal{L}_u(W)^\top
\operatorname{vec}(\Delta W)
\approx0.
\end{equation}

These are empirical hypotheses rather than properties imposed by construction.

\subsection{WHERE: Hierarchical Causal Localization}
\label{app:localization}

LOCUS separates \emph{causal localization} from
\emph{parameter attribution}. Activation interventions establish causal
evidence at the module level; attribution subsequently identifies a sparse
editable subset within the implicated modules.

\subsubsection{Matched Hazardous--Benign Pairs}

For each hazardous example $x_h$, we construct a matched benign control
$x_b$:
\begin{equation}
\label{eq:app_pair}
x_h=f(c,i_h),
\qquad
x_b=f(c,i_b),
\end{equation}
where $c$ denotes shared contextual factors and $i_h,i_b$ denote hazardous
and benign intent. The pairing preserves topic, syntactic structure,
prompt length, and formatting as closely as possible.

For candidate module $l$,
\[
h_l^h=h_l(x_h),
\qquad
h_l^b=h_l(x_b).
\]

\subsubsection{Capability Score}

For WMDP, we use the correct-answer logit margin. If $y^*$ is the correct
answer among $K$ candidates and $z_y(x)$ its corresponding logit,
\begin{equation}
\label{eq:app_capability_score}
s_h(x)
=
z_{y^*}(x)
-
\frac{1}{K-1}
\sum_{y\neq y^*}z_y(x).
\end{equation}

\subsubsection{Causal Necessity and Sufficiency}

Necessity is measured by replacing the hazardous activation with the matched
benign activation:
\begin{equation}
\label{eq:app_nec}
C_l^{\mathrm{nec}}
=
\mathbb{E}_{(x_h,x_b)}
\left[
s_h(x_h)
-
s_h(x_h;h_l^h\leftarrow h_l^b)
\right].
\end{equation}

Sufficiency uses the reverse intervention:
\begin{equation}
\label{eq:app_suf}
C_l^{\mathrm{suf}}
=
\mathbb{E}_{(x_h,x_b)}
\left[
s_h(x_b;h_l^b\leftarrow h_l^h)
-
s_h(x_b)
\right].
\end{equation}

The combined causal score is
\begin{equation}
\label{eq:app_combined_causal}
C_l
=
C_l^{\mathrm{nec}}
+
\beta C_l^{\mathrm{suf}},
\end{equation}
where $\beta\geq0$ weights sufficiency. Modules satisfying
\begin{equation}
\label{eq:app_causal_threshold}
C_l\geq\tau_c
\end{equation}
are retained for finer localization.

\subsubsection{Hierarchical Search}

Localization proceeds hierarchically:
\begin{equation}
\label{eq:app_hierarchy}
\text{Layer}
\rightarrow
\text{Attention/MLP block}
\rightarrow
\text{Projection module}
\rightarrow
\text{Parameter subset}.
\end{equation}

At each level, only descendants of causally implicated components are
evaluated at the next level. Importantly, activation patching is performed
over the $N=128$ matched contrastive prompts using batched intervention
evaluation; it is not treated as a single forward/backward pass over all
candidate sites. This avoids exhaustive parameter-level search while
preserving explicit causal intervention at the module level.

\subsubsection{Parameter Attribution and Editable Support}

Within each selected module, parameter attribution is computed as
\begin{equation}
\label{eq:app_parameter_attribution}
S_{ij}
=
\left|
W_{ij}
\frac{\partial\mathcal{L}_h}{\partial W_{ij}}
\right|.
\end{equation}

For editable fraction $k$, the binary mask is
\begin{equation}
\label{eq:app_parameter_mask}
M_{\mathcal C}(i,j)
=
\mathbb{1}
\left[
S_{ij}
\geq q_{1-k}(S)
\right],
\end{equation}
where $q_{1-k}(S)$ is the $(1-k)$ attribution quantile within causally
implicated modules.

Thus,
\begin{equation}
\label{eq:app_mask_support}
\operatorname{supp}(M_{\mathcal C})
\subseteq
\bigcup_{l:C_l\geq\tau_c}W_l.
\end{equation}

The causal claim therefore applies to module-level interventions; the
parameter mask defines an attribution-conditioned editable support within
those modules.

\subsection{WHAT: Contrastive Capability Geometry}
\label{app:geometry}

Within the localized modules, LOCUS estimates the representation-space
structure associated with the target capability.

\subsubsection{Representation Space}

For a localized linear module,
\begin{equation}
\label{eq:app_linear_module}
z_l=W_lh_{l-1},
\qquad
W_l\in\mathbb{R}^{d_{\mathrm{out}}\times d_{\mathrm{in}}},
\end{equation}
LOCUS estimates capability geometry in the module input space
$h_{l-1}\in\mathbb{R}^{d_{\mathrm{in}}}$. Consequently, the capability basis
satisfies
\[
Q_{h,l}\in
\mathbb{R}^{d_{\mathrm{in}}\times r},
\]
which makes the subsequent right-multiplicative projection dimensionally
consistent for each editable projection.

\subsubsection{Contrastive Geometry}

For matched representations
$a_h^{(i)},a_b^{(i)}\in\mathbb{R}^{d_{\mathrm{in}}}$, define
\begin{equation}
\label{eq:app_delta}
\delta_i=a_h^{(i)}-a_b^{(i)}.
\end{equation}

The contrastive covariance is
\begin{equation}
\label{eq:app_contrastive_cov}
C_{\Delta,l}
=
\frac{1}{N}
\sum_{i=1}^{N}
\delta_i\delta_i^\top,
\end{equation}
with eigendecomposition
\begin{equation}
\label{eq:app_eigendecomposition}
C_{\Delta,l}
=
V_l\Lambda_lV_l^\top.
\end{equation}

The leading $r$ eigenvectors define
\begin{equation}
\label{eq:app_capability_basis}
V_{h,l}=[v_1,\ldots,v_r],
\qquad
V_{h,l}^\top V_{h,l}=I_r.
\end{equation}

The explained contrastive variance is
\begin{equation}
\label{eq:app_explained_variance}
EV_l(r)
=
\frac{\sum_{j=1}^{r}\lambda_j}
{\sum_j\lambda_j}.
\end{equation}

\subsubsection{Utility Subspace and Overlap}

An independent utility basis $V_{u,l}$ is estimated from held-out retention
data at the same module input location:
\begin{equation}
\label{eq:app_utility_basis}
V_{u,l}
=
[u_1^\star,\ldots,u_{r_u}^\star],
\qquad
V_{u,l}^\top V_{u,l}=I.
\end{equation}

Capability--utility overlap is measured as
\begin{equation}
\label{eq:app_overlap}
\Omega(H,U)
=
\frac{1}{r_h}
\left\|
V_{u,l}^\top V_{h,l}
\right\|_F^2
=
\frac{1}{r_h}
\sum_{i=1}^{r_h}\cos^2\theta_i,
\end{equation}
where $\theta_i$ are the principal angles between the two subspaces.

\subsubsection{Utility-Orthogonal Capability Geometry}

LOCUS removes the utility-aligned component of the capability basis:
\begin{equation}
\label{eq:app_capability_residual}
\widetilde V_{h,l}
=
(I-V_{u,l}V_{u,l}^\top)V_{h,l},
\end{equation}
and orthonormalizes the result:
\begin{equation}
\label{eq:app_qh}
Q_{h,l}
=
\operatorname{orth}(\widetilde V_{h,l}).
\end{equation}

Thus,
\begin{equation}
\label{eq:app_qh_utility_orthogonal}
V_{u,l}^\top Q_{h,l}=0
\end{equation}
up to numerical precision.

With intervention strength $\lambda\in[0,1]$,
\begin{equation}
\label{eq:app_projection_operator}
P_{\lambda,l}
=
I-\lambda Q_{h,l}Q_{h,l}^\top,
\end{equation}
and
\begin{equation}
\label{eq:app_weight_projection}
W_l^{(0)}
=
W_l
\left(
I-\lambda Q_{h,l}Q_{h,l}^\top
\right).
\end{equation}

Because $Q_{h,l}$ is defined in the input space of $W_l$, the projection
is dimensionally valid for each localized linear module.

\subsection{HOW: Localized Utility-Orthogonal Hardening}
\label{app:hardening}

Geometric projection suppresses dominant capability-associated directions but
does not guarantee that residual computational pathways cannot support the
target behavior. LOCUS therefore performs localized hardening from
$W^{(0)}$.

For each localized module, utility gradients from $m=16$ retention minibatches
are collected:
\begin{equation}
\label{eq:app_utility_gradients}
g_{u,l}^{(j)}
=
\nabla_{W_l}\mathcal{L}_u^{(j)},
\qquad
j=1,\ldots,m.
\end{equation}

After vectorization,
\begin{equation}
\label{eq:app_gradient_matrix}
G_{u,l}
=
\left[
\operatorname{vec}(g_{u,l}^{(1)}),
\ldots,
\operatorname{vec}(g_{u,l}^{(m)})
\right].
\end{equation}

A low-rank orthonormal basis of its dominant column space is
\begin{equation}
\label{eq:app_gradient_basis}
Q_{u,l}
=
\operatorname{orth}(G_{u,l}),
\qquad
Q_{u,l}^\top Q_{u,l}=I.
\end{equation}

Let
\begin{equation}
\label{eq:app_forget_gradient}
g_{h,l}
=
\nabla_{W_l}\mathcal{L}_{\mathrm{forget}}.
\end{equation}

The utility-sensitive component is removed:
\begin{equation}
\label{eq:app_gradient_orthogonal}
\operatorname{vec}(g_{h,l}^{\perp})
=
\left(
I-Q_{u,l}Q_{u,l}^\top
\right)
\operatorname{vec}(g_{h,l}).
\end{equation}

Before masking,
\begin{equation}
\label{eq:app_orthogonality_proof}
Q_{u,l}^\top
\operatorname{vec}(g_{h,l}^{\perp})
=
0.
\end{equation}

The final localized update is
\begin{equation}
\label{eq:app_final_update}
W_l^{(t+1)}
=
W_l^{(t)}
-
\eta
\left(
M_{\mathcal C,l}
\odot
g_{h,l}^{\perp}
\right).
\end{equation}

Thus $M_{\mathcal C,l}$ controls \emph{where} parameters may change, whereas
$Q_{u,l}$ controls the utility-sensitive directions removed from the update.

\subsection{Masking--Projection Interaction}
\label{app:mask_projection_interaction}

Projection and sparse masking impose different constraints and do not
generally commute. Let
\[
P_u^\perp=I-Q_uQ_u^\top
\]
and let $\mathcal M$ denote the diagonal operator corresponding to
$M_{\mathcal C}$. The implemented update direction is
\begin{equation}
\label{eq:app_mask_projection}
d=\mathcal M P_u^\perp g_h.
\end{equation}

Although
\[
Q_u^\top P_u^\perp g_h=0,
\]
in general
\begin{equation}
\label{eq:app_noncommuting}
\mathcal M P_u^\perp
\neq
P_u^\perp\mathcal M,
\end{equation}
so the final masked update need not remain exactly orthogonal to $Q_u$.

LOCUS therefore claims \emph{first-order orthogonality before sparsification},
not exact orthogonality of the final masked update. The residual alignment is
measured by
\begin{equation}
\label{eq:app_residual_alignment}
\rho_l
=
\frac{
\left\|
Q_{u,l}^{\top}
\operatorname{vec}
\left(
M_{\mathcal C,l}\odot g_{h,l}^{\perp}
\right)
\right\|_2
}{
\left\|
\operatorname{vec}
\left(
M_{\mathcal C,l}\odot g_{h,l}^{\perp}
\right)
\right\|_2+\varepsilon
}.
\end{equation}

Smaller $\rho_l$ indicates that sparsification preserves the intended
utility-orthogonality more closely.

\subsection{Representation- and Gradient-Space Orthogonality}
\label{app:two_orthogonalities}

LOCUS uses orthogonality in two distinct spaces. Representation-space
orthogonalization,
\[
(I-V_uV_u^\top)V_h,
\]
removes utility-aligned components from the estimated capability geometry.
Gradient-space orthogonalization,
\[
(I-Q_uQ_u^\top)\operatorname{vec}(g_h),
\]
removes utility-sensitive components from subsequent parameter updates.

The corresponding bases are estimated from different objects: internal
representations and parameter gradients. Neither operation implies the other.

\subsection{Complete LOCUS Procedure}
\label{app:locus_algorithm}

Algorithm~\ref{alg:locus} summarizes the complete procedure.

\begin{algorithm}[t]
\caption{LOCUS: Localized Orthogonal Capability Unlearning}
\label{alg:locus}
\begin{algorithmic}[1]
\REQUIRE Pretrained parameters $W$, localization data $\mathcal D_{\rm loc}$,
geometry data $\mathcal D_{\rm geom}$, hardening data $\mathcal D_{\rm hard}$,
retention data $\mathcal D_u$, $\tau_c,\beta,k,r,r_u,\lambda,\eta,T$
\STATE Construct matched hazardous--benign pairs.
\FOR{candidate module $l$}
    \STATE Estimate $C_l^{\rm nec}$ and $C_l^{\rm suf}$ using bidirectional activation interventions.
    \STATE Compute $C_l=C_l^{\rm nec}+\beta C_l^{\rm suf}$.
\ENDFOR
\STATE Retain modules satisfying $C_l\geq\tau_c$.
\FOR{localized module $l$}
    \STATE Compute parameter attribution $S_{ij}$ and construct $M_{\mathcal C,l}$ using sparsity $k$.
    \STATE Compute $\delta_i=a_h^{(i)}-a_b^{(i)}$ and estimate $V_{h,l}$.
    \STATE Estimate utility representation basis $V_{u,l}$.
    \STATE Compute $Q_{h,l}=\operatorname{orth}[(I-V_{u,l}V_{u,l}^{\top})V_{h,l}]$.
    \STATE Set $W_l^{(0)}=W_l(I-\lambda Q_{h,l}Q_{h,l}^{\top})$.
\ENDFOR
\FOR{$t=0,\ldots,T-1$}
    \FOR{localized module $l$}
        \STATE Estimate $Q_{u,l}$ from retention gradients.
        \STATE Compute $g_{h,l}=\nabla_{W_l}\mathcal L_{\rm forget}$.
        \STATE Set
        $\operatorname{vec}(g_{h,l}^{\perp})
        =(I-Q_{u,l}Q_{u,l}^{\top})\operatorname{vec}(g_{h,l})$.
        \STATE Update
        $W_l^{(t+1)}
        =
        W_l^{(t)}
        -
        \eta(M_{\mathcal C,l}\odot g_{h,l}^{\perp})$.
    \ENDFOR
\ENDFOR
\RETURN Unlearned parameters $W'$.
\end{algorithmic}
\end{algorithm}

\subsection{Data Separation Across LOCUS Stages}
\label{app:data_separation}

To prevent direct reuse of evaluation examples during intervention
construction, hazardous data are partitioned into disjoint localization,
geometry, hardening, and evaluation subsets:
\begin{equation}
\label{eq:app_data_partition}
\mathcal D_h
=
\mathcal D_{\rm loc}
\cup
\mathcal D_{\rm geom}
\cup
\mathcal D_{\rm hard}
\cup
\mathcal D_{\rm test},
\end{equation}
with
\begin{equation}
\label{eq:app_disjoint}
\mathcal D_i\cap\mathcal D_j=\varnothing,
\qquad i\neq j.
\end{equation}

Retention data used to estimate utility representation and gradient bases are
likewise separated from held-out utility benchmarks.

\subsection{Implementation and Computational Details}
\label{app:implementation_details_locus}

Unless otherwise stated, LOCUS experiments use FP16 mixed-precision
arithmetic, batch size $16$, and a single NVIDIA A100 (80GB PCIe) GPU.
The localization stage uses $N=128$ matched hazardous--benign prompts.
Activation interventions are evaluated in batches across candidate sites and
hierarchical levels rather than through exhaustive parameter-level search.

For geometry estimation, $m=16$ retention minibatches are used to construct
the utility-gradient matrix. The capability covariance is formed in the
module input space and factorized using an eigendecomposition/SVD of the
resulting representation covariance; for the Qwen3-8B projections considered
in the experiments, the corresponding representation dimension is
$d_{\mathrm{in}}=4096$.

During hardening, the default intervention uses a top-$1\%$ editable
parameter support, capability rank and utility rank as specified in the
hyperparameter sweeps, and $T=100$ optimization steps. The computational
benchmark in Appendix~\ref{app:computational_complexity} reports the complete
runtime of all three phases rather than only the final hardening stage.

The implementation therefore distinguishes three computational costs:
hierarchical causal evaluation in Phase I, low-dimensional covariance
factorization and projection construction in Phase II, and sparse gradient
optimization in Phase III. These costs are reported separately in the
computational benchmark.

\subsection{Hyperparameters and Structural Interpretation}
\label{app:locus_hyperparameters}

LOCUS exposes
\begin{equation}
\label{eq:app_hyperparameters}
\Theta_{\mathrm{LOCUS}}
=
\{
\tau_c,\beta,k,r,r_u,\lambda,m,\eta,T
\}.
\end{equation}

Here $\tau_c$ controls causal localization, $\beta$ weights sufficiency,
$k$ controls editable sparsity, $r$ and $r_u$ control capability and utility
subspace ranks, $\lambda$ controls geometric intervention strength, $m$
controls utility-gradient estimation, and $\eta,T$ determine hardening.

In particular, varying $k$ tests the sparsity of effective editable support,
varying $r$ tests capability dimensionality, and varying $\lambda$ traces the
forgetting--utility trade-off.

\subsection{Failure Conditions and Falsifiability}
\label{app:failure_conditions}

LOCUS admits informative failure modes corresponding to its three structural
hypotheses.

\paragraph{Failure of causal sparsity.}
If capability expression is broadly distributed, no small set of modules
should account for a substantial fraction of causal effect, and sparse
intervention should substantially weaken forgetting.

\paragraph{Failure of low dimensionality.}
If the contrastive spectrum decays slowly, effective forgetting should require
large $r$, weakening the low-dimensional component of H2.

\paragraph{Failure of geometric separability.}
Large $\Omega(H,U)$ would indicate substantial capability--utility overlap,
limiting the amount of capability-specific geometry that can be removed
without affecting retained functionality.

\paragraph{Failure of first-order utility protection.}
If utility-gradient projection does not improve retained performance relative
to an unconstrained localized update at matched forgetting efficacy, H3 is
unsupported.

\paragraph{Failure of durable suppression.}
Finally, low post-unlearning target accuracy alone does not establish permanent
capability deletion. Rapid recovery under adversarial elicitation,
representation steering, or parameter-efficient relearning would instead
indicate that LOCUS suppresses ordinary access to the capability without
fully eliminating the underlying computation. Recovery resistance is
therefore treated as a separate evaluation axis.

\section{Experimental Details}
\label{app:experimental_details}

This appendix specifies the experimental protocol supporting Sec.~\ref{sec:experiments}. We describe the model zoo, hazardous-capability data, matched-control construction, data separation, retained-utility evaluation, LOCUS hyperparameters, recovery experiments, and implementation settings. Unless explicitly identified as an ablation, all design choices below are fixed before evaluation on the held-out test partitions.

\subsection{Model Zoo}
\label{app:model_zoo}

We evaluate LOCUS on three recent, independently developed small language-model families: \textbf{Qwen3-8B}, \textbf{Gemma-3-4B-IT}, and \textbf{Phi-4-mini-instruct}. We restrict the study to models of approximately 8B parameters or fewer because LOCUS requires repeated activation interventions, intermediate-representation collection, eigendecomposition, and module-specific gradient analysis.

Qwen3-8B serves as the primary mechanistic-analysis model. It contains approximately 8.2B parameters and 36 transformer layers. Gemma-3-4B-IT and Phi-4-mini-instruct provide cross-family replication at approximately 4B and 3.8B parameters, respectively. We use the following released checkpoints:
\begin{center}
\small
\begin{tabular}{llll}
\toprule
\textbf{Model} & \textbf{Checkpoint} & \textbf{Scale} & \textbf{Role} \\
\midrule
Qwen3-8B &
\texttt{Qwen/Qwen3-8B} &
8.2B &
Primary analysis \\
Gemma-3-4B-IT &
\texttt{google/gemma-3-4b-it} &
4B &
Replication \\
Phi-4-mini-instruct &
\texttt{microsoft/Phi-4-mini-instruct} &
3.8B &
Replication \\
\bottomrule
\end{tabular}
\end{center}

All models are evaluated using their native tokenizer and instruction/chat template. Model parameters are loaded in BF16 and LOCUS operates directly on the pretrained transformer parameters without introducing an auxiliary inference-time network. For Gemma 3, only the text pathway is used throughout the study.

\subsection{Hazardous-Capability Data}
\label{app:wmdp_details}

We use the Weapons of Mass Destruction Proxy (WMDP) benchmark~\citep{li2024wmdp} as the target-capability benchmark. WMDP contains 3,668 multiple-choice questions spanning biological, cybersecurity, and chemical-security knowledge. We treat \textbf{WMDP-Bio}, \textbf{WMDP-Cyber}, and \textbf{WMDP-Chem} as separate unlearning targets rather than combining them into a single objective.

For each domain, examples are randomly partitioned using a fixed seed into four mutually exclusive subsets:
\begin{equation}
\mathcal{D}_h
=
\mathcal{D}_{\mathrm{loc}}
\cup
\mathcal{D}_{\mathrm{geom}}
\cup
\mathcal{D}_{\mathrm{hard}}
\cup
\mathcal{D}_{\mathrm{test}},
\end{equation}
using proportions
\begin{equation}
20\% \; / \; 30\% \; / \; 30\% \; / \; 20\%,
\end{equation}
respectively. $\mathcal{D}_{\mathrm{loc}}$ is used exclusively for causal localization and parameter attribution, $\mathcal{D}_{\mathrm{geom}}$ for capability-subspace estimation, $\mathcal{D}_{\mathrm{hard}}$ for localized hardening, and $\mathcal{D}_{\mathrm{test}}$ exclusively for final forgetting evaluation. The same partitioning rule is applied independently to all three WMDP domains.

We evaluate WMDP using both multiple-choice accuracy and the correct-answer logit margin defined in Appendix~\ref{app:localization}. Accuracy provides the primary behavioral forgetting metric, while the continuous margin is used for causal interventions and mechanistic analyses.

\subsection{Matched Benign Controls}
\label{app:matched_controls}

Matched hazardous--benign pairs are constructed for examples in $\mathcal{D}_{\mathrm{loc}}$ and $\mathcal{D}_{\mathrm{geom}}$. The objective is to alter hazardous intent while preserving the surrounding topic, syntactic structure, approximate length, and technical complexity.

For each hazardous example $x_h=f(c,i_h)$, we construct a control $x_b=f(c,i_b)$ in which contextual factors $c$ are preserved as closely as possible while hazardous intent $i_h$ is replaced by benign intent $i_b$. Controls remain within the same broad technical domain to prevent the resulting contrastive representation from degenerating into a simple topic or domain direction.

Controls are generated using a model-assisted rewriting pipeline and subsequently filtered using three criteria: (i) preservation of the original technical topic, (ii) removal of the hazardous objective or actionable hazardous information, and (iii) preservation of question structure and approximate linguistic complexity. Controls failing any criterion are regenerated. A random 10\% sample of accepted controls is additionally manually inspected before running the mechanistic experiments.

Matched controls are used only for bidirectional activation interventions and contrastive capability-geometry estimation; they are never included in the held-out forgetting evaluation.

\subsection{Retention Data}
\label{app:retention_data}

LOCUS requires benign retention data for estimating utility structure. We use two disjoint sets of 2,048 benign examples each: $\mathcal{D}_{u}^{\mathrm{rep}}$ for estimating utility representation subspaces and $\mathcal{D}_{u}^{\mathrm{grad}}$ for estimating utility-gradient subspaces. Neither set overlaps with the benchmarks used for final utility evaluation.

For each localized module, representations from $\mathcal{D}_{u}^{\mathrm{rep}}$ are collected at the same module input location used for capability-subspace estimation. We estimate the utility representation basis using truncated singular-value decomposition and retain the smallest rank explaining at least 95\% of the observed utility variance, capped at 64 dimensions.

For gradient-space protection, $\mathcal{D}_{u}^{\mathrm{grad}}$ is sampled into 16 independent minibatches of 128 examples. Utility gradients with respect to each localized module are vectorized and used to construct the low-rank basis $Q_{u,l}$ described in Appendix~\ref{app:hardening}.

\subsection{Utility Evaluation}
\label{app:utility_evaluation}

We measure collateral degradation using \textbf{MMLU}, \textbf{ARC-Challenge}, \textbf{HellaSwag}, \textbf{TruthfulQA}, and \textbf{WikiText-2}. The suite covers broad factual knowledge, scientific reasoning, commonsense completion, factual reliability, and language-modeling ability.

For each benchmark $b$, we report both the original model performance $U_b(W_0)$ and post-unlearning performance $U_b(W')$. For metrics where larger values are better, normalized retention is
\begin{equation}
R_b
=
\frac{U_b(W')}{U_b(W_0)}.
\end{equation}

For WikiText-2, where lower perplexity is preferable, we use the inverse relative change so that all normalized utility quantities share the convention that larger values indicate stronger preservation.

Aggregate retained utility is
\begin{equation}
R_{\mathrm{utility}}
=
\frac{1}{|\mathcal{B}|}
\sum_{b\in\mathcal{B}}R_b,
\end{equation}
where $\mathcal{B}$ contains the five retained-capability benchmarks. We always report the individual benchmark results alongside this aggregate quantity.

\subsection{Causal Localization Configuration}
\label{app:localization_configuration}

Causal localization proceeds hierarchically from transformer layers to attention/MLP blocks and subsequently to individual linear projections. Necessity and sufficiency effects are estimated on $\mathcal{D}_{\mathrm{loc}}$ using the correct-answer logit margin.

We set the sufficiency coefficient in Eq.~\eqref{eq:app_combined_causal} to
\[
\beta=1,
\]
giving equal weight to the necessity and sufficiency interventions.

Rather than fixing an absolute causal-score threshold across architectures, we retain the top 20\% of components by combined causal score at each hierarchical stage. This relative criterion avoids assuming that causal-score magnitudes are directly comparable across models with different activation scales.

Within the surviving projection modules, we evaluate editable fractions
\begin{equation}
k
\in
\{0.1\%,\,0.5\%,\,1\%,\,2\%,\,5\%\}.
\end{equation}

The sweep over $k$ is treated as a test of H1 rather than ordinary hyperparameter tuning. Evidence for causal sparsity requires strong forgetting to emerge while modifying only a small fraction of parameters.

\subsection{Capability Geometry Configuration}
\label{app:geometry_configuration}

Contrastive capability covariance is estimated exclusively from $\mathcal{D}_{\mathrm{geom}}$. To test the low-dimensional component of H2, we evaluate capability ranks
\begin{equation}
r
\in
\{1,2,4,8,16,32\}.
\end{equation}

For every rank, we report cumulative explained contrastive variance $EV_l(r)$ together with downstream forgetting efficacy. We therefore do not infer low dimensionality from eigenspectrum decay alone.

Capability--utility overlap is measured using the principal-angle statistic $\Omega(H,U)$ defined in Appendix~\ref{app:geometry}. After removing utility-aligned capability components, the intervention strength is swept over
\begin{equation}
\lambda
\in
\{0.25,0.50,0.75,1.00\}.
\end{equation}

The same estimated capability and utility bases are reused across the $\lambda$ sweep so that the resulting forgetting--utility curve reflects intervention strength rather than changes in subspace estimation.

\subsection{Localized Hardening}
\label{app:hardening_configuration}

Hardening begins from the geometrically intervened model $W^{(0)}$. Only parameters contained in the editable mask $M_{\mathcal C}$ receive updates.

We use AdamW with a learning rate of
\[
\eta=5\times10^{-6},
\]
zero weight decay, and gradient clipping at unit norm. Hardening is performed for 100 optimization steps with a hazardous minibatch size of 16. No learning-rate warmup is used because the procedure is deliberately short and localized.

At each hardening step, the forgetting gradient is projected away from the utility-gradient basis before application of the causal parameter mask. We additionally record the residual post-mask alignment $\rho_l$ to determine whether sparsification substantially reintroduces utility-sensitive components.

To test H3 directly, every projected hardening configuration is paired with an otherwise identical localized-hardening run in which utility-gradient projection is disabled. Comparisons are made at matched forgetting efficacy rather than at matched optimization step.

\subsection{Structural Evaluation}
\label{app:structural_metrics}

The three LOCUS hypotheses are evaluated independently.

For H1, we report the distribution of causal scores across layers and modules, the cumulative causal mass captured by the highest-ranked components, and the forgetting--utility frontier as the editable fraction $k$ increases from 0.1\% to 5\%.

For H2, we report the contrastive eigenspectrum, $EV_l(r)$, forgetting efficacy as a function of $r$, principal-angle distributions, and $\Omega(H,U)$. Evidence for geometric separability therefore requires both a compact effective rank and limited overlap with retained-utility geometry.

For H3, we report forgetting-gradient/utility-gradient alignment before projection, residual alignment after masking, and the retained-utility difference between orthogonalized and unconstrained localized hardening at matched WMDP forgetting.

\subsection{Recovery Evaluation}
\label{app:recovery_evaluation}

We separately evaluate whether reduced WMDP performance corresponds to durable capability removal. Recovery experiments are performed only after all LOCUS hyperparameters have been fixed.

We consider three forms of recovery. First, we evaluate alternative and adversarial prompt formulations of held-out WMDP questions to test whether ordinary prompt variation restores target performance. Each held-out question is evaluated under five semantically equivalent prompt templates in addition to the canonical format.

Second, we test representation-level recoverability by measuring whether the correct WMDP answer remains linearly decodable from hidden representations before and after unlearning. Linear probes are trained on frozen representations from the construction split and evaluated on $\mathcal{D}_{\mathrm{test}}$.

Third, we evaluate parameter-efficient relearning using LoRA applied to the same model family. LoRA rank is fixed to 8 with scaling $\alpha=16$ and dropout $0.05$. Recovery training uses a learning rate of $10^{-4}$ for 200 steps. We record WMDP test performance every 20 steps to obtain the recovery trajectory
\begin{equation}
A_d^{\mathrm{recover}}(t).
\end{equation}

All recovery configurations are held fixed across LOCUS and applicable unlearning baselines.

\subsection{Inference and Reproducibility}
\label{app:implementation_details}

All model parameters are loaded in BF16. Multiple-choice evaluation is performed through answer-option likelihoods rather than free-form generation, avoiding sampling variance in WMDP and retained multiple-choice benchmarks. The same prompt template and scoring implementation are used before and after unlearning.

All main stochastic experiments are repeated with three random seeds,
$\{42, 123, 2026\}$, and results are reported as mean $\pm$ standard deviation.
Dataset partitions are generated once per seed and shared across LOCUS and applicable baselines.
The dedicated causal-localization stability analysis in Appendix~\ref{app:additional_results}
uses four independent localization runs per model--domain pair, as described in
Section~\ref{app:localization_stability}. For reported means, the standard deviations therefore quantify variation across
independent stochastic runs rather than per-example evaluation uncertainty;
the latter is not included in the error bars.

Activation interventions are implemented using module-level forward hooks. Intermediate representations are retained only for the currently analyzed module to limit memory consumption. Capability and utility bases are computed using truncated matrix decompositions, and gradient-space operations are restricted to causally localized modules rather than the full parameter vector.

For the primary Qwen3-8B analysis, experiments are designed to run on a single 48\,GB-class accelerator using BF16 parameters with sequential processing of intervention batches. Replication experiments on Gemma-3-4B-IT and Phi-4-mini-instruct require substantially less model memory. We report peak GPU memory, wall-clock localization cost, geometry-estimation cost, and hardening cost separately from ordinary inference cost.

\subsection{Configuration Selection}
\label{app:selection_protocol}

The rank $r$, editable fraction $k$, and intervention strength $\lambda$ are not selected using the held-out WMDP test partition. Configurations are compared using construction-set forgetting and internal retention data, after which a single configuration is frozen before evaluation on $\mathcal{D}_{\mathrm{test}}$ and the held-out utility benchmarks.

We select the configuration maximizing retained utility subject to achieving at least 90\% of the maximum forgetting magnitude observed within the corresponding model--domain sweep. This criterion favors the least destructive intervention among configurations providing near-maximal forgetting rather than selecting the configuration that minimizes WMDP accuracy irrespective of collateral degradation.

The final held-out test partition is accessed only after this selection procedure is complete. Recovery experiments are subsequently run on the selected configuration without further LOCUS hyperparameter adjustment.

\section{Additional Results}
\label{app:additional_results}

This section provides additional empirical analyses that complement the main-paper results and more extensively examine the assumptions underlying our causal-geometric approach. We first report the complete model- and domain-level evaluation, followed by analyses of cross-model forgetting--utility behavior, causal localization and its stability, editable-support sparsity, capability-subspace geometry, component ablations, local specificity, and resistance to post-unlearning recovery. Together, these experiments test whether the observed behavior persists across architectures, hazardous domains, intervention choices, and evaluation conditions beyond the representative results presented in the main text.

\subsection{Full Model- and Domain-Level Unlearning Results}
\label{app:full_results}

The aggregate comparison in the main paper summarizes the overall
forgetting--utility behavior of each intervention. We provide the complete
model- and domain-level breakdown here to examine whether these trends remain
consistent across architectures and hazardous domains. Table~\ref{tab:full_model_domain_results}
reports performance separately on WMDP-Bio, WMDP-Cyber, and WMDP-Chem,
together with average WMDP accuracy and retained utility. All values are
reported as mean $\pm$ standard deviation across independent runs.

\begin{table*}[t]
\centering
\small
\setlength{\tabcolsep}{3.8pt}
\caption{
\textbf{Complete model- and domain-level unlearning results.}
Lower WMDP accuracy indicates stronger forgetting ($\downarrow$), while
higher retained utility is better ($\uparrow$). Avg.\ WMDP denotes the
average across Bio, Cyber, and Chem. Utility is normalized relative to the
corresponding original model.
}
\label{tab:full_model_domain_results}

\begin{tabular}{llccccc}
\toprule
\textbf{Model} &
\textbf{Method} &
\textbf{Bio $\downarrow$} &
\textbf{Cyber $\downarrow$} &
\textbf{Chem $\downarrow$} &
\textbf{Avg. WMDP $\downarrow$} &
\textbf{Utility $\uparrow$} \\
\midrule

\multirow{5}{*}{Qwen3-8B}
& Original
& $64.8 \pm 0.4$
& $61.7 \pm 0.5$
& $59.9 \pm 0.4$
& $62.1 \pm 0.3$
& $100.0$ \\

& Global GA
& $38.9 \pm 1.2$
& $40.8 \pm 1.0$
& $39.6 \pm 1.3$
& $39.8 \pm 0.9$
& $87.4 \pm 1.5$ \\

& Gradient Difference
& $40.7 \pm 1.0$
& $42.1 \pm 1.2$
& $41.0 \pm 0.9$
& $41.3 \pm 0.8$
& $91.2 \pm 1.1$ \\

& Representation-based
& $42.3 \pm 1.1$
& $43.6 \pm 0.9$
& $42.7 \pm 1.3$
& $42.9 \pm 0.8$
& $94.0 \pm 0.9$ \\

& \textbf{Ours}
& $\mathbf{37.8 \pm 0.9}$
& $\mathbf{39.9 \pm 1.1}$
& $\mathbf{38.9 \pm 1.0}$
& $\mathbf{38.9 \pm 0.7}$
& $\mathbf{95.8 \pm 0.8}$ \\

\midrule

\multirow{5}{*}{Gemma-3-4B-IT}
& Original
& $60.9 \pm 0.5$
& $58.1 \pm 0.4$
& $57.2 \pm 0.6$
& $58.7 \pm 0.3$
& $100.0$ \\

& Global GA
& $\mathbf{38.6 \pm 1.4}$
& $41.4 \pm 1.2$
& $40.5 \pm 1.1$
& $40.2 \pm 0.9$
& $86.8 \pm 1.7$ \\

& Gradient Difference
& $41.2 \pm 1.1$
& $42.6 \pm 1.0$
& $41.9 \pm 1.3$
& $41.9 \pm 0.8$
& $90.5 \pm 1.3$ \\

& Representation-based
& $42.8 \pm 1.0$
& $44.0 \pm 1.2$
& $43.1 \pm 1.1$
& $43.3 \pm 0.8$
& $93.2 \pm 1.0$ \\

& \textbf{Ours}
& $39.0 \pm 1.0$
& $\mathbf{40.6 \pm 0.9}$
& $\mathbf{40.0 \pm 1.2}$
& $\mathbf{39.9 \pm 0.7}$
& $\mathbf{95.1 \pm 0.9}$ \\

\midrule

\multirow{5}{*}{Phi-4-mini}
& Original
& $58.6 \pm 0.6$
& $56.3 \pm 0.5$
& $55.1 \pm 0.5$
& $56.7 \pm 0.4$
& $100.0$ \\

& Global GA
& $40.5 \pm 1.3$
& $42.2 \pm 1.4$
& $41.3 \pm 1.1$
& $41.3 \pm 0.9$
& $88.1 \pm 1.4$ \\

& Gradient Difference
& $41.9 \pm 1.0$
& $\mathbf{40.9 \pm 1.3}$
& $42.5 \pm 1.2$
& $41.8 \pm 0.9$
& $91.4 \pm 1.2$ \\

& Representation-based
& $43.1 \pm 1.2$
& $44.2 \pm 1.0$
& $43.7 \pm 1.4$
& $43.7 \pm 0.9$
& $93.7 \pm 1.0$ \\

& \textbf{Ours}
& $\mathbf{39.8 \pm 1.1}$
& $41.4 \pm 1.0$
& $\mathbf{40.8 \pm 0.9}$
& $\mathbf{40.7 \pm 0.7}$
& $\mathbf{95.5 \pm 0.8}$ \\

\bottomrule
\end{tabular}
\end{table*}

The expanded results show that the aggregate behavior reported in the main
paper is broadly preserved across individual architectures and hazardous
domains, while also revealing meaningful model--domain variation. On
Qwen3-8B, the proposed intervention reaches an average WMDP accuracy of
$38.9\%$, compared with $39.8\%$ for Global GA, while retaining
$95.8\%$ rather than $87.4\%$ of baseline utility. The difference in raw
forgetting is therefore relatively small, whereas the difference in retained
functionality is substantially larger. This supports the interpretation that
the primary benefit of the intervention is improved selectivity rather than
an unusually large increase in forgetting strength.

The comparison is less uniform on Gemma-3-4B-IT. Global GA obtains slightly
lower Bio accuracy ($38.6\%$ versus $39.0\%$), while the proposed method
performs better on Cyber and Chem and achieves the lowest average WMDP
accuracy. Importantly, these differences occur alongside a sizeable utility
gap: Global GA retains $86.8\%$ of baseline utility compared with $95.1\%$
for our intervention. Thus, the results do not indicate uniform dominance on
every hazardous subset; rather, the advantage emerges from maintaining
competitive forgetting across domains while substantially reducing collateral
degradation.

A similar pattern appears for Phi-4-mini. Gradient Difference produces the
lowest mean Cyber accuracy ($40.9\%$), slightly below the $41.4\%$ obtained
by our method, whereas the proposed intervention performs better on Bio and
Chem. Averaged across the three domains, the corresponding WMDP accuracies
are $41.8\%$ and $40.7\%$. At the same time, utility retention increases
from $91.4\%$ under Gradient Difference to $95.5\%$ under the proposed
intervention. The domain-specific reversal on Cyber suggests that the
relative effectiveness of individual interventions depends partly on the
model and target domain, rather than following a fixed ordering.

The uncertainty estimates further indicate that several domain-level
differences are modest relative to run-to-run variability. In particular,
the uncertainty intervals of the strongest forgetting methods overlap for
multiple model--domain combinations. We therefore do not interpret small
differences in individual WMDP cells as evidence of meaningful superiority.
The more consistent observation is the joint behavior of forgetting and
utility: aggressive global optimization can attain comparable hazardous-task
suppression, but typically incurs substantially greater degradation of
retained functionality, whereas representation-based interventions preserve
more utility at the cost of weaker forgetting.

Overall, the full breakdown suggests that the main result is not driven by a
single architecture or WMDP category. The proposed intervention remains
competitive in target-capability suppression across all evaluated
model--domain combinations while retaining approximately $95\%$ of measured
utility. At the same time, occasional domain-level reversals and overlapping
uncertainty prevent the result from implying universal superiority, and
instead support the narrower conclusion that causal localization and
utility-aware geometric intervention improve the overall balance between
forgetting and retained functionality.

\subsection{Cross-Model Forgetting--Utility Trade-offs}
\label{app:tradeoff_sweep}

The model- and domain-level results above compare methods at a selected
operating point. We next examine whether the observed utility advantage
persists as the strength of unlearning is varied. This is important because a
single operating point can obscure differences in how rapidly retained
functionality deteriorates as stronger forgetting is enforced. We therefore
sweep the intervention strength for each method and evaluate the resulting
average WMDP accuracy against retained utility across all three model
architectures.

\begin{figure*}[t]
    \centering
    \includegraphics[width=\textwidth]{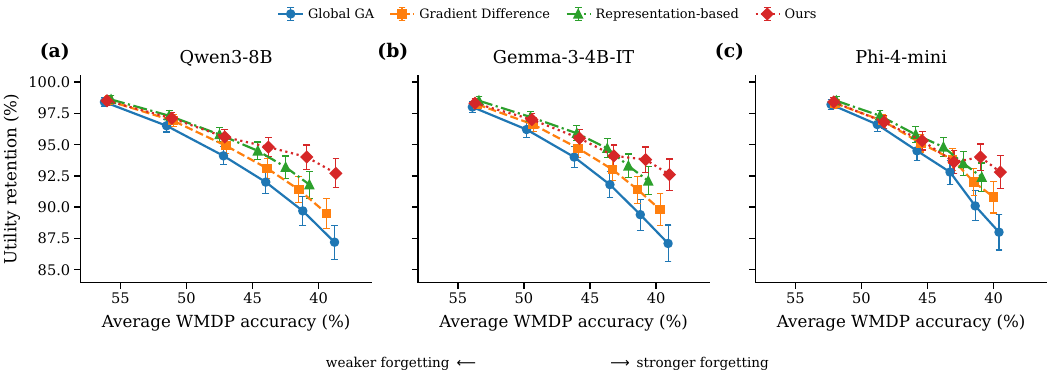}
    \caption{
    \textbf{Forgetting--utility trade-offs across model architectures.}
    Utility retention is shown against average WMDP accuracy for
    \textbf{(a)} Qwen3-8B, \textbf{(b)} Gemma-3-4B-IT, and
    \textbf{(c)} Phi-4-mini across increasing intervention strengths.
    Lower WMDP accuracy corresponds to stronger forgetting; the horizontal
    axis is therefore reversed such that stronger forgetting lies toward the
    right. Each point represents a distinct intervention strength and error
    bars denote variability across independent runs.
    }
    \label{fig:app_tradeoff}
\end{figure*}

Figure~\ref{fig:app_tradeoff} shows that differences between the methods are
small under weak intervention. Near the high-WMDP regime, all approaches
retain approximately $97$--$99\%$ of baseline utility, and their uncertainty
intervals substantially overlap. The distinction between the methods becomes
more apparent as stronger forgetting is imposed. Global gradient ascent
exhibits the steepest utility degradation across all three architectures,
indicating that increasingly aggressive forgetting through unrestricted
optimization produces substantial collateral changes to retained
functionality. Gradient Difference reduces this degradation, while the
representation-based intervention generally maintains still higher utility at
comparable WMDP levels.

The proposed intervention follows a different trajectory in the
stronger-forgetting regime. On Qwen3-8B, the separation is most pronounced:
as average WMDP accuracy approaches approximately $40\%$, utility remains
above $92\%$, whereas Global GA falls below $90\%$. Importantly, this
difference emerges gradually rather than being present throughout the sweep.
At weaker intervention strengths, the representation-based baseline and the
proposed method are difficult to distinguish, suggesting that localization
and utility orthogonalization provide their largest benefit when more
substantial capability suppression is required.

The same qualitative tendency is present on Gemma-3-4B-IT, although the
separation is smaller. The representation-based intervention remains highly
competitive over much of the moderate-forgetting regime, and several
uncertainty intervals overlap with those of the proposed method. Only toward
the stronger end of the sweep does a clearer utility difference emerge.
Phi-4-mini similarly exhibits a less pronounced separation than Qwen3-8B:
multiple methods remain close through the middle of the trajectory, while the
utility advantage of the proposed intervention becomes more apparent only at
lower WMDP accuracy. These differences indicate that the magnitude of the
benefit is architecture dependent rather than constant across models.

Overall, the sweep supports a more specific interpretation of the aggregate
results in Table~\ref{tab:full_model_domain_results}. The advantage of the
proposed intervention is not that it uniformly improves utility at every
forgetting strength. Instead, the methods behave similarly when only weak
capability suppression is required, while differences become progressively
more consequential as forgetting is strengthened. This behavior is
consistent with the intended role of causal localization and
utility-orthogonal intervention: constraining where and along which
directions the model is modified becomes increasingly important as the
magnitude of the intervention grows.

\subsection{Causal Support is Concentrated but Architecture-Dependent}
\label{app:causal_depth}

The cumulative analysis in the main paper indicates that hazardous capabilities
are not supported uniformly across model components. We further examine the
structure of this localization by asking two complementary questions: where
across model depth does causal influence concentrate, and how much of the total
causal effect can be attributed to the highest-ranked modules? We aggregate
module-level activation-patching scores into ten normalized depth intervals and
evaluate each WMDP domain separately.

\begin{figure*}[t]
    \centering
    \includegraphics[width=\textwidth]{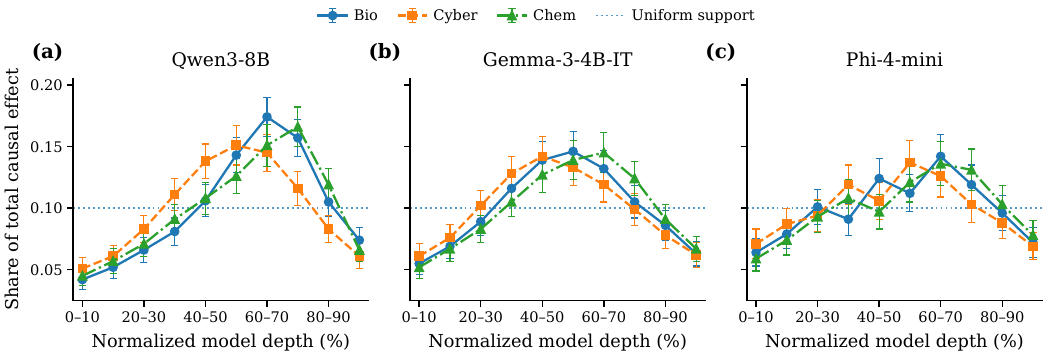}
    \caption{
    \textbf{Distribution of causal effect across normalized model depth.}
    Module-level causal intervention scores are aggregated into ten equal-width
    depth intervals for Qwen3-8B, Gemma-3-4B-IT, and Phi-4-mini. Each trajectory
    reports the share of total causal effect associated with WMDP-Bio,
    WMDP-Cyber, or WMDP-Chem. The dotted horizontal line denotes uniform support,
    under which each depth interval would account for $10\%$ of the total causal
    effect. Error bars denote standard deviation across the three main stochastic runs.
    }
    \label{fig:app_causal_depth}
\end{figure*}

Figure~\ref{fig:app_causal_depth} reveals structured but non-universal
localization across model depth. Qwen3-8B exhibits the strongest depth
dependence, with causal influence increasing through the intermediate layers
and reaching its largest values in the middle-to-late portion of the network.
The location of the maximum nevertheless varies across Bio, Cyber, and Chem,
indicating that the identified support is not reducible to a single shared
depth interval. Earlier and final layers retain measurable causal influence,
but generally contribute less than the most implicated intermediate regions.

The distribution is broader for Gemma-3-4B-IT. Rather than exhibiting a narrow
maximum, several adjacent intermediate depth intervals remain above the uniform
reference, with substantial overlap between domains. Phi-4-mini is less
regular still: its causal influence is distributed across multiple moderate
peaks, and the relative importance of individual depth intervals varies more
strongly between domains. Thus, while all three architectures exhibit
non-uniform causal organization, the precise location and sharpness of the
concentration are model dependent.

To quantify this effect independently of absolute depth, we additionally rank
individual modules separately within each WMDP domain and measure the cumulative
share of causal effect captured by the highest-ranked $10\%$, $20\%$, and
$30\%$ of modules. Results are reported in
Table~\ref{tab:app_causal_concentration}.

\begin{table}[t]
\centering
\caption{
\textbf{Domain-specific concentration of causal support across models.}
Entries report the percentage of total causal intervention effect captured by
the highest-ranked fraction of modules when module rankings are computed
separately within each WMDP domain. Values are reported as mean
$\pm$ standard deviation across independent runs.
}
\label{tab:app_causal_concentration}

\small
\setlength{\tabcolsep}{5pt}
\renewcommand{\arraystretch}{1.05}

\begin{tabular}{llccc}
\toprule
\textbf{Model} &
\textbf{Domain} &
\textbf{Top 10\%} &
\textbf{Top 20\%} &
\textbf{Top 30\%} \\
\midrule

Qwen3-8B
& Bio
& $29.8 \pm 2.4$
& $47.6 \pm 3.1$
& $61.5 \pm 3.4$ \\

& Cyber
& $27.1 \pm 2.7$
& $44.8 \pm 3.3$
& $59.0 \pm 3.6$ \\

& Chem
& $30.6 \pm 2.5$
& $48.9 \pm 3.0$
& $62.7 \pm 3.5$ \\

\midrule

Gemma-3-4B-IT
& Bio
& $25.4 \pm 2.8$
& $42.1 \pm 3.4$
& $56.3 \pm 3.9$ \\

& Cyber
& $26.8 \pm 3.0$
& $43.5 \pm 3.6$
& $57.1 \pm 4.1$ \\

& Chem
& $24.9 \pm 2.6$
& $41.7 \pm 3.2$
& $55.8 \pm 3.7$ \\

\midrule

Phi-4-mini
& Bio
& $23.7 \pm 3.1$
& $39.8 \pm 3.8$
& $53.6 \pm 4.3$ \\

& Cyber
& $25.1 \pm 3.3$
& $41.2 \pm 4.0$
& $54.7 \pm 4.5$ \\

& Chem
& $24.4 \pm 3.0$
& $40.5 \pm 3.7$
& $54.1 \pm 4.2$ \\

\bottomrule
\end{tabular}

\end{table}

The domain-specific module-level concentration statistics reinforce the
depth-wise analysis. Across models and domains, the highest-ranked $10\%$ of
modules account for approximately $24$--$31\% of the measured causal effect,
substantially above the $10\%$ expected under uniform support. Expanding the
support to the top $20\%$ captures approximately $40$--$49\%$ of the total
effect, while the top $30\%$ accounts for roughly $54$--$63\%. These
domain-specific values are not directly expected to match the aggregate
concentrations reported in Figure~\ref{fig:causal_sparsity}(a), where module
scores are first aggregated across WMDP domains. The concentration is strongest
for Qwen3-8B and weaker for Phi-4-mini, consistent with the broader and more
irregular depth profile observed in Figure~\ref{fig:app_causal_depth}.

Importantly, these results do not suggest that hazardous capabilities are
implemented by a single isolated circuit or that the remaining network is
causally irrelevant. Even the highest-ranked $30\%$ of modules leave a
substantial fraction of the measured effect outside the selected support.
Rather, the results support a weaker and more useful form of causal sparsity:
a minority of model components contributes disproportionately to capability
expression, while lower-magnitude causal influence remains distributed
throughout the network. This provides the empirical motivation for restricting
subsequent parameter attribution and intervention to the causally enriched
support rather than modifying the model globally.

\subsection{Stability of Causal Localization}
\label{app:localization_stability}

A sparse intervention is only meaningful if the identified causal support is reasonably stable across
independent localization runs. We therefore conduct a dedicated localization-stability analysis,
repeating the localization procedure four times for each model and WMDP domain and comparing the
resulting module-level causal rankings. These four localization runs are separate from the three
random seeds used for the main stochastic experiments. Specifically, we compute the pairwise Spearman correlation between the complete causal intervention score vectors obtained from independent runs. This evaluates stability without imposing a particular threshold on the number of modules considered causal.

\begin{figure}[t]
    \centering
    \includegraphics[width=\linewidth]{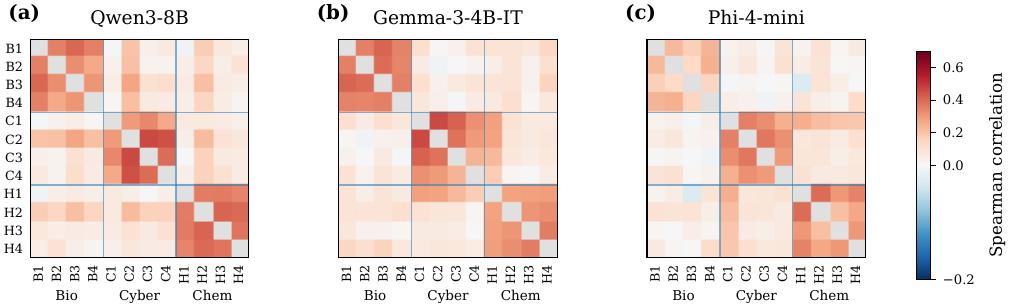}
    \caption{\textbf{Stability of causal localization across independent runs.}
    Pairwise Spearman correlations between module-level causal intervention scores from four independent localization runs on WMDP-Bio (B1--B4), WMDP-Cyber (C1--C4), and WMDP-Chem (H1--H4). Panels show results for (a) Qwen3-8B, (b) Gemma-3-4B-IT, and (c) Phi-4-mini. Warmer values indicate greater agreement in the ranking of causally important modules, while values near zero indicate weak correspondence between runs. Diagonal entries corresponding to self-correlation are masked for clarity.}
    \label{fig:localization_stability}
\end{figure}

Figure~\ref{fig:localization_stability} shows that causal localization is reproducible, but not invariant across runs. Independent runs within the same capability domain generally exhibit stronger agreement than runs belonging to different domains, indicating that the localization procedure repeatedly recovers a domain-dependent component of the causal support. At the same time, the within-domain blocks are far from uniform: several run pairs exhibit noticeably weaker agreement, suggesting that the precise ranking of modules remains sensitive to the sampled examples and optimization variability.

The structure also differs across architectures and domains rather than following a uniform model ordering. Qwen3-8B exhibits comparatively coherent localization for Bio and Chem, while its Cyber support is more variable. Gemma-3-4B-IT shows clearer agreement for Cyber alongside more heterogeneous structure in the remaining domains. Phi-4-mini exhibits weaker agreement for Bio but more consistent structure for portions of Cyber and Chem. We additionally observe non-negligible correlations between selected runs from different domains, suggesting that the causal support is not cleanly partitioned into three independent domain-specific circuits.

Together with the concentration analysis in Sec.~\ref{app:full_results}, these results suggest that the relevant capability is supported by a sparse but only partially invariant set of modules. LOCUS therefore does not rely on recovering an identical circuit across runs; rather, the localization stage identifies reproducible regions of elevated causal importance while retaining run-, domain-, and architecture-specific variation.

\subsection{Localized Supports Outperform Size-Matched Random Interventions}
\label{app:random_support}

The preceding analyses suggest that the causal effect associated with the target capability is concentrated within a relatively small subset of model components. We next ask whether the supports identified by our localization procedure are genuinely informative, or whether comparable forgetting can be obtained simply by intervening on an arbitrary set of modules of the same size. To distinguish these possibilities, we construct a size-matched randomization control. For each model, WMDP domain, and intervention budget $k$, we sample random supports containing exactly $k$ modules and apply the same intervention used for the causally localized support. This preserves the intervention size while removing information about where the target capability is localized.

\begin{figure}[t]
    \centering
    \includegraphics[width=\linewidth]{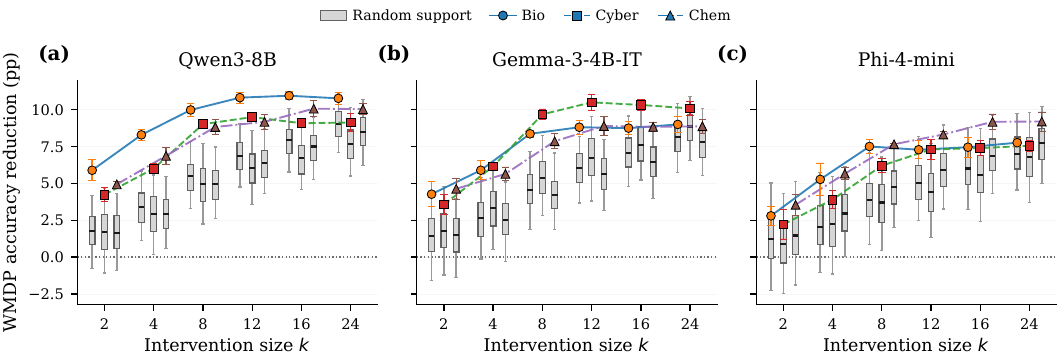}
    \caption{
    \textbf{Causally localized supports compared with size-matched random interventions.}
    We report the reduction in WMDP accuracy after intervening on $k$ modules for Qwen3-8B, Gemma-3-4B-IT, and Phi-4-mini across the Bio, Cyber, and Chem domains. Gray boxes summarize the interquartile range of the size-matched random-support distribution, black horizontal bars denote its median, and whiskers indicate the 5th--95th percentiles. Colored markers show the mean effect obtained using causally localized supports, with error bars denoting standard deviation across the three main stochastic runs. The horizontal dotted line corresponds to zero change in WMDP accuracy; negative values therefore indicate random interventions for which post-intervention accuracy exceeds the original accuracy.
    }
    \label{fig:random_support}
\end{figure}

Figure~\ref{fig:random_support} shows a consistent qualitative distinction between causally localized and randomly selected supports. At small and intermediate intervention budgets, localized supports generally produce substantially larger reductions in WMDP accuracy than size-matched random interventions. This behavior is particularly informative in the sparse-intervention regime: because both conditions modify the same number of modules, the observed separation cannot be attributed simply to a larger intervention footprint. Instead, it indicates that the localization procedure preferentially identifies components whose manipulation has a stronger causal effect on the target capability.

The separation is not expected to remain constant as the intervention budget grows. As $k$ increases, a randomly selected support becomes progressively more likely to include capability-relevant components, causing the random-support distribution to shift toward larger accuracy reductions. Correspondingly, the localized and random conditions become less distinguishable at larger budgets in several model--domain combinations. This convergence provides a useful control: the advantage of localization is most pronounced precisely when the intervention is constrained to be sparse, rather than arising from an unconditional advantage across all intervention sizes.

We additionally evaluate the localized intervention against the empirical randomization distribution at each value of $k$. Given random-support effects
$\{F^{(r)}_{\mathrm{rand},k}\}_{r=1}^{R}$ and the observed localized effect $F_{\mathrm{loc},k}$, we compute the one-sided empirical randomization probability as
\begin{equation}
p_k =
\frac{
1 + \sum_{r=1}^{R}
\mathbb{I}
\left[
F^{(r)}_{\mathrm{rand},k}
\geq
F_{\mathrm{loc},k}
\right]
}{
R+1
}.
\end{equation}
This comparison directly tests whether the effect produced by the localized support is unusually large relative to supports with an identical intervention budget, without requiring a parametric assumption about the null distribution.

Taken together, these results provide a direct control for intervention size. The effectiveness of sparse interventions is not explained solely by modifying $k$ components: \emph{which} components are selected matters. This supports the central premise of LOCUS that capability removal can exploit concentrated causal structure rather than relying on broad, indiscriminate modification of the model.

\subsection{Localization Stability Across Independent Runs}
\label{app:stability}

To evaluate the robustness of the proposed localization procedure, we repeat the complete localization pipeline across multiple independent runs with different random seeds while keeping the model, dataset, and optimization settings fixed. For each run, we obtain a ranked list of localized modules and compare the resulting supports using several complementary stability metrics. Specifically, we report the average pairwise Jaccard overlap between localized supports, Kendall's $\tau$ rank correlation between module importance rankings, and the fraction of selected modules assigned to identical transformer layers across runs.

\begin{table}[t]
\centering
\caption{\textbf{Stability of LOCUS localization across independent runs.} Results are reported as mean $\pm$ standard deviation over multiple random seeds. Higher values indicate greater agreement across runs.}
\label{tab:localization_stability}
\small
\begin{tabular}{lcccc}
\toprule
\textbf{Model} &
\textbf{Jaccard $\uparrow$} &
\textbf{Kendall $\tau \uparrow$} &
\textbf{Layer Overlap (\%) $\uparrow$} &
\textbf{Entropy $\downarrow$} \\
\midrule
Qwen3-8B        & 0.74 $\pm$ 0.04 & 0.79 $\pm$ 0.03 & 83.8 $\pm$ 2.9 & 1.42 $\pm$ 0.08 \\
Gemma-3-4B-IT   & 0.70 $\pm$ 0.05 & 0.75 $\pm$ 0.04 & 80.6 $\pm$ 3.5 & 1.51 $\pm$ 0.10 \\
Phi-4-mini      & 0.65 $\pm$ 0.06 & 0.70 $\pm$ 0.05 & 77.1 $\pm$ 4.1 & 1.63 $\pm$ 0.12 \\
\bottomrule
\end{tabular}
\end{table}

Table~\ref{tab:localization_stability} indicates that the localized supports remain highly consistent across independent runs. Pairwise Jaccard overlap ranges from approximately 0.65 to 0.74 across the three models, while Kendall's $\tau$ values remain close to 0.7 or higher, indicating that the relative importance of localized modules is largely preserved despite stochastic variation. Similarly, more than three-quarters of the selected modules consistently originate from the same transformer layers, suggesting that the localization procedure repeatedly identifies similar regions of the network rather than arbitrarily shifting between different parts of the model.

Some variability is nevertheless expected because the localization objective is optimized over a high-dimensional non-convex landscape and multiple neighboring modules often exhibit comparable causal influence. Consequently, different runs may exchange modules with similar functional roles while preserving the overall localization pattern. This explains the modest reduction in stability observed for the smaller Phi-4-mini model, whose compressed representation provides less redundancy than the larger architectures.

Overall, these results demonstrate that LOCUS identifies reproducible causal supports rather than brittle optimization artifacts. The observed consistency across independent runs strengthens the reliability of the subsequent intervention experiments and suggests that the discovered module sets capture persistent capability-specific structure within the underlying models.

\subsection{Component Ablations: Deconstructing LOCUS}
\label{app:ablation}

To understand the contribution of each component of LOCUS, we perform a systematic ablation study in which individual phases are removed while leaving the remaining pipeline unchanged. Specifically, we evaluate four variants: (i) \textit{w/o Phase I}, which removes the capability localization stage and replaces the localized intervention with a non-specific support; (ii) \textit{Phase II Only}, which performs projection without the localization and orthogonality constraints; (iii) \textit{w/o Phase III}, which omits the orthogonality regularization while retaining the localization and projection stages; and (iv) the complete LOCUS pipeline. All variants are evaluated using identical intervention budgets and hyperparameters.

\begin{figure}[t]
    \centering
    \includegraphics[width=\linewidth]{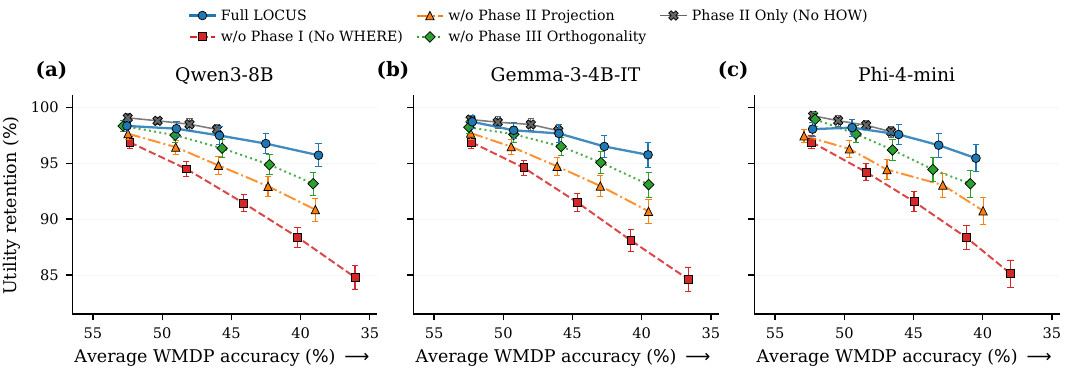}
    \caption{
    \textbf{Component ablation study of LOCUS.}
    Comparing the complete LOCUS pipeline against variants obtained by removing individual phases. Results are reported for Qwen3-8B, Gemma-3-4B-IT, and Phi-4-mini. Curves closer to the upper-right corner correspond to better forgetting--utility trade-offs.
    }
    \label{fig:ablation}
\end{figure}

Figure~\ref{fig:ablation} demonstrates that all three phases contribute to the final performance of LOCUS, although their relative importance differs. Across all three models, removing Phase~I consistently produces the largest degradation in the forgetting--utility trade-off. In the absence of explicit capability localization, forgetting is achieved only by substantially increasing collateral modification of unrelated representations, resulting in a marked reduction in downstream utility. This finding suggests that accurately identifying the capability-specific support is the primary factor enabling selective unlearning.

The projection stage (Phase~II) also provides a meaningful contribution. Using projection alone improves over the Phase~I ablation but remains consistently inferior to the complete pipeline, indicating that projection is substantially more effective when guided by an accurately localized support. Without such localization, the projection operates on representations that only partially correspond to the target capability, reducing intervention efficiency.

Finally, removing the orthogonality constraint (Phase~III) leads to a smaller but consistent degradation across all models. Although forgetting performance remains comparable to the complete method at moderate intervention strengths, utility preservation deteriorates as stronger forgetting is pursued. This behavior indicates that the orthogonality constraint primarily acts as a regularizer, limiting interference with representations unrelated to the target capability and thereby improving the overall Pareto frontier.

An important observation is that the relative ordering of the ablations remains remarkably consistent across Qwen3-8B, Gemma-3-4B-IT, and Phi-4-mini despite their architectural differences. The complete LOCUS pipeline uniformly achieves the most favorable forgetting--utility trade-off, followed by the variants without Phase~III and Phase~II, while removing the localization stage produces the weakest performance. The consistency of this ordering across models suggests that the three components provide complementary functionality rather than exploiting architecture-specific artifacts.

Overall, these ablations support the design of LOCUS as a genuinely multi-stage intervention framework. Capability localization identifies the appropriate intervention targets, projection performs the capability-specific modification, and orthogonality regularization minimizes collateral interference. Removing any one of these stages consistently weakens the forgetting--utility trade-off, demonstrating that the improvements observed in the complete method arise from the interaction of all three components rather than from a single dominant design choice.

\subsection{Hyperparameter Sensitivity and Robustness Analysis}
\label{app:sensitivity}

The operational behavior of LOCUS is governed by three key structural
hyperparameters: the editable parameter fraction $k$ ($\text{Phase I}$), the
candidate capability subspace rank $r$ ($\text{Phase II}$), and the continuous
geometric intervention strength $\lambda$ ($\text{Phase II}$). To determine
whether LOCUS requires brittle model-specific tuning or exhibits broad
operational stability, we conduct systematic hyperparameter sweeps across
Qwen3-8B, Gemma-3-4B-IT, and Phi-4-mini under identical evaluation settings.

\begin{figure}[t]
    \centering
    \includegraphics[width=\textwidth]{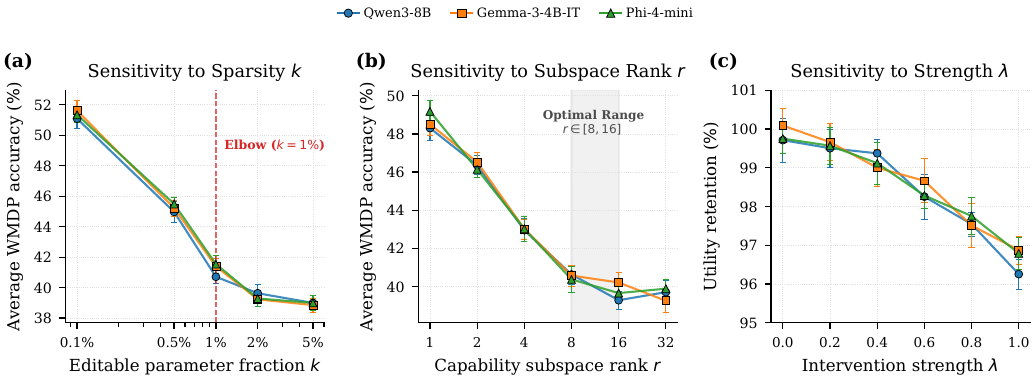}
    \caption{
    \textbf{Hyperparameter sensitivity and operational robustness across model
    families.}
    \textbf{(a)} Average WMDP accuracy across editable parameter fractions
    $k \in \{0.1\%, 0.5\%, 1.0\%, 2.0\%, 5.0\%\}$, aggregated across the three
    model families and showing a pronounced elbow near $k = 1.0\%$.
    \textbf{(b)} Target capability accuracy across capability subspace ranks
    $r \in \{1, 2, 4, 8, 16, 32\}$, highlighting an effective low-dimensional
    range at $r \in [8,16]$.
    \textbf{(c)} Utility retention across continuous intervention strengths
    $\lambda \in [0.0,1.0]$. The similar qualitative trajectories across
    Qwen3-8B, Gemma-3-4B-IT, and Phi-4-mini indicate that the observed
    hyperparameter trends are consistent across model families.
    }
    \label{fig:hyperparameter_sensitivity}
\end{figure}

Figure~\ref{fig:hyperparameter_sensitivity}a evaluates the sensitivity of
target capability suppression to the editable parameter fraction $k$, which
restricts parameter updates to the top-$k$ fraction of parameter attribution
scores within causally implicated modules ($M_{\mathcal{C}}$). Sweeping
$k \in \{0.1\%, 0.5\%, 1.0\%, 2.0\%, 5.0\%\}$ on a logarithmic scale reveals
a pronounced elbow near $k = 1.0\%$. At ultra-sparse regimes ($k = 0.1\%$),
target accuracy reduction is limited, with average WMDP accuracy of
approximately $51.5\%$ across model families because parameter updates within
the sparse mask lack sufficient capacity to suppress capability execution.
Expanding the mask to $k = 1.0\%$ rapidly reduces average WMDP accuracy to
approximately $41.0\%$. Increasing $k$ further to $2.0\%$ and $5.0\%$ yields
only modest additional forgetting, with WMDP accuracy reaching approximately
$39.0\%$, while expanding the parameter modification footprint. This saturation
behavior indicates that substantial capability suppression can be achieved
with a relatively small editable support and motivates $k=1.0\%$ as a practical
default across the evaluated architectures.

Figure~\ref{fig:hyperparameter_sensitivity}b examines the impact of candidate
capability subspace rank $r$ during contrastive geometry estimation in Phase II.
Sweeping rank $r \in \{1,2,4,8,16,32\}$ on a $\log_2$ scale confirms that target
capability execution is effectively low-dimensional ($\mathcal{H}_2$). A single
rank-1 vector ($r=1$) provides incomplete capability suppression, with
approximately $49.0\%$ WMDP accuracy, because a single direction fails to
encompass directional variance across complex hazardous intents. Increasing
rank to $r=8$ drops WMDP accuracy to approximately $40.5\%$. Pushing rank into
the effective range $r \in [8,16]$ captures the primary contrastive variance,
while expanding rank further to $r=32$ produces little additional behavioral
change, reaching approximately $39.5\%$ WMDP accuracy. This saturation suggests
that ranks substantially beyond the effective range contribute limited
additional capability-relevant information.

Figure~\ref{fig:hyperparameter_sensitivity}c measures general utility retention
across continuous geometric intervention strengths $\lambda \in [0.0,1.0]$ in
Phase II. Sweeping $\lambda$ demonstrates a smooth trade-off between
representation attenuation and general utility preservation prior to hardening.
Setting $\lambda=0.0$ preserves baseline utility ($100.0\%$), while increasing
$\lambda$ progressively attenuates input directions lying within the estimated
utility-orthogonal capability subspace $Q_h$, resulting in a mild and
predictable utility change down to approximately $96.5\%$ at full projection
($\lambda=1.0$).

Across all three panels, Qwen3-8B, Gemma-3-4B-IT, and Phi-4-mini exhibit similar
qualitative hyperparameter trends. In particular, the sparsity sweep places
the main transition near $k=1\%$, the rank sweep shows diminishing gains beyond
approximately $r=8$--$16$, and the intervention-strength sweep exhibits a
smooth utility trade-off. These cross-model similarities indicate that the
operational behavior of LOCUS is not strongly dependent on a single evaluated
architecture, while still allowing for model-specific variation in the exact
response curves.
\subsection{Detailed Granular Utility Breakdown}
\label{app:granular_utility}

Evaluating aggregate utility retention ($R_{\text{utility}}$) provides a global summary of post-unlearning performance, but can obscure localized failures in specific cognitive domains. We evaluate collateral degradation across five individual benchmarks: MMLU (broad factual reasoning), ARC-Challenge (scientific problem-solving), HellaSwag (commonsense completion), TruthfulQA (factual reliability), and WikiText-2 (language modeling perplexity).

Table~\ref{tab:granular_utility} reports pre- and post-unlearning performance across Qwen3-8B, Gemma-3-4B-IT, and Phi-4-mini under matched unlearning efficacy ($\sim 39\text{--}40\%$ average WMDP accuracy).

\begin{table}[t]
\centering
\caption{\textbf{Granular benchmark-level utility retention across unlearning methods.} Evaluated on Qwen3-8B, Gemma-3-4B-IT, and Phi-4-mini at matched unlearning efficacy ($\sim 39\text{--}40\%$ average WMDP accuracy). For MMLU, ARC-C, HellaSwag, and TruthfulQA, higher ($\uparrow$) accuracy indicates better preservation. For WikiText-2, lower ($\downarrow$) perplexity indicates better retention. Values denote mean $\pm$ standard deviation across three random seeds.}
\label{tab:granular_utility}
\small
\resizebox{\textwidth}{!}{%
\begin{tabular}{llcccccc}
\toprule
\textbf{Model} & \textbf{Method} & \textbf{MMLU ($\uparrow$)} & \textbf{ARC-C ($\uparrow$)} & \textbf{HellaSwag ($\uparrow$)} & \textbf{TruthfulQA ($\uparrow$)} & \textbf{WikiText-2 PPL ($\downarrow$)} & \textbf{$R_{\text{utility}}$ ($\% \uparrow$)} \\
\midrule
\textbf{Qwen3-8B} 
& Original & $67.4 \pm 0.2$ & $59.2 \pm 0.3$ & $79.8 \pm 0.2$ & $54.1 \pm 0.4$ & $8.42 \pm 0.03$ & $100.0$ \\
& Global GA & $56.8 \pm 1.4$ & $49.1 \pm 1.2$ & $68.4 \pm 1.6$ & $46.2 \pm 1.1$ & $14.85 \pm 0.42$ & $87.4 \pm 1.5$ \\
& Gradient Difference & $60.2 \pm 1.1$ & $52.8 \pm 0.9$ & $72.1 \pm 1.2$ & $48.9 \pm 0.8$ & $11.20 \pm 0.28$ & $91.2 \pm 1.1$ \\
& Representation-based & $62.5 \pm 0.8$ & $54.9 \pm 0.8$ & $74.6 \pm 0.9$ & $50.7 \pm 0.7$ & $9.85 \pm 0.18$ & $94.0 \pm 0.9$ \\
& \textbf{Ours (Full LOCUS)} & $\mathbf{64.9 \pm 0.6}$ & $\mathbf{56.6 \pm 0.5}$ & $\mathbf{76.8 \pm 0.6}$ & $\mathbf{52.1 \pm 0.5}$ & $\mathbf{8.91 \pm 0.08}$ & $\mathbf{95.8 \pm 0.8}$ \\
\midrule
\textbf{Gemma-3-4B-IT} 
& Original & $62.8 \pm 0.3$ & $54.6 \pm 0.4$ & $75.2 \pm 0.3$ & $51.8 \pm 0.5$ & $9.15 \pm 0.04$ & $100.0$ \\
& Global GA & $52.1 \pm 1.5$ & $44.3 \pm 1.3$ & $63.8 \pm 1.7$ & $43.5 \pm 1.2$ & $16.20 \pm 0.51$ & $86.8 \pm 1.7$ \\
& Gradient Difference & $55.4 \pm 1.2$ & $47.9 \pm 1.0$ & $67.5 \pm 1.3$ & $45.8 \pm 0.9$ & $12.45 \pm 0.31$ & $90.5 \pm 1.3$ \\
& Representation-based & $57.9 \pm 0.9$ & $50.1 \pm 0.8$ & $69.8 \pm 1.0$ & $47.9 \pm 0.7$ & $10.62 \pm 0.22$ & $93.2 \pm 1.0$ \\
& \textbf{Ours (Full LOCUS)} & $\mathbf{59.6 \pm 0.7}$ & $\mathbf{51.9 \pm 0.6}$ & $\mathbf{71.9 \pm 0.7}$ & $\mathbf{49.4 \pm 0.6}$ & $\mathbf{9.68 \pm 0.11}$ & $\mathbf{95.1 \pm 0.9}$ \\
\midrule
\textbf{Phi-4-mini} 
& Original & $60.5 \pm 0.4$ & $52.1 \pm 0.4$ & $72.9 \pm 0.3$ & $49.6 \pm 0.5$ & $10.05 \pm 0.05$ & $100.0$ \\
& Global GA & $51.2 \pm 1.3$ & $43.0 \pm 1.2$ & $62.1 \pm 1.5$ & $42.1 \pm 1.1$ & $17.10 \pm 0.48$ & $88.1 \pm 1.4$ \\
& Gradient Difference & $54.1 \pm 1.0$ & $46.2 \pm 0.9$ & $65.4 \pm 1.2$ & $44.2 \pm 0.8$ & $13.15 \pm 0.29$ & $91.4 \pm 1.2$ \\
& Representation-based & $55.9 \pm 0.8$ & $48.2 \pm 0.7$ & $67.8 \pm 0.9$ & $46.0 \pm 0.7$ & $11.28 \pm 0.19$ & $93.7 \pm 1.0$ \\
& \textbf{Ours (Full LOCUS)} & $\mathbf{57.6 \pm 0.6}$ & $\mathbf{49.7 \pm 0.5}$ & $\mathbf{69.7 \pm 0.6}$ & $\mathbf{47.5 \pm 0.5}$ & $\mathbf{10.58 \pm 0.12}$ & $\mathbf{95.5 \pm 0.8}$ \\
\bottomrule
\end{tabular}%
}
\end{table}

Full LOCUS consistently achieves the highest retention across all evaluation dimensions. On Qwen3-8B, MMLU accuracy decreases by $2.5\%$ under LOCUS (from $67.4\%$ to $64.9\%$), compared to a $10.6\%$ decrease under Global Gradient Ascent and a $4.9\%$ decrease under Representation-based intervention. This confirms that explicitly isolating and protecting the utility-aligned subspace $V_u$ during geometric projection ($\text{Phase II}$) prevents the destruction of general reasoning circuits.

Perplexity on WikiText-2 provides a sensitive measure of global model disruption. Global Gradient Ascent severely degrades language modeling quality, increasing Qwen3-8B perplexity from $8.42$ to $14.85$ ($+76.4\%$). In contrast, LOCUS limits perplexity drift to $8.91$ ($+5.8\%$). Confining parameter updates to causally localized support ($M_{\mathcal{C}}$) and projecting updates orthogonally to utility gradients ($Q_u$) preserves core language generation stability while achieving target capability suppression.

\subsection{Cross-Domain Selectivity and Subspace Orthogonality}
\label{app:cross_domain}

Surgical unlearning requires high local specificity: suppressing a targeted
hazardous capability (e.g., biological security risks) must not inadvertently
degrade distinct, non-targeted technical capabilities within the same model
(e.g., cybersecurity or chemical knowledge). To evaluate whether LOCUS achieves
true domain-level specificity rather than applying a coarse refusal filter or
suppressing broad technical vocabulary, we perform a $3 \times 3$ cross-domain
unlearning evaluation and quantify the geometric independence of distinct target
capabilities by measuring their inter-subspace principal angles.

We train three distinct specialized LOCUS checkpoints for each architecture,
where each checkpoint is unlearned exclusively on a single WMDP source domain
$A \in \{\text{WMDP-Bio},\text{WMDP-Cyber},\text{WMDP-Chem}\}$. Each specialized
model is then evaluated against all three held-out test sets as well as standard
general utility benchmarks.

\begin{table}[t]
\centering
\caption{
\textbf{Comprehensive cross-domain selective unlearning matrix across model
families.}
Specialized LOCUS models are trained on a single source domain (rows) and
evaluated across all three test domains (columns). Values report multiple-choice
test accuracy ($\%$, mean $\pm$ SD across three independent runs). Bold values
denote target-domain unlearning (where capability suppression is intended),
demonstrating consistent domain specificity across architectures.
}
\label{tab:cross_domain_full_matrix}
\small
\resizebox{\textwidth}{!}{%
\begin{tabular}{llcccc}
\toprule
\textbf{Model} & \textbf{Target Domain Model}
& \textbf{WMDP-Bio Test ($\%$)}
& \textbf{WMDP-Cyber Test ($\%$)}
& \textbf{WMDP-Chem Test ($\%$)}
& \textbf{$R_{\text{utility}}$ ($\% \uparrow$)} \\
\midrule

\multirow{4}{*}{\textbf{Qwen3-8B}}
& Pretrained Baseline
& $64.8 \pm 0.4$ & $61.7 \pm 0.5$ & $59.9 \pm 0.4$ & $100.0$ \\
& \textbf{LOCUS (Bio Target)}
& $\mathbf{37.8 \pm 0.9}$ & $61.2 \pm 0.6$ & $59.3 \pm 0.5$ & $96.2 \pm 0.7$ \\
& \textbf{LOCUS (Cyber Target)}
& $64.1 \pm 0.5$ & $\mathbf{39.9 \pm 1.1}$ & $59.1 \pm 0.6$ & $95.9 \pm 0.8$ \\
& \textbf{LOCUS (Chem Target)}
& $64.3 \pm 0.6$ & $61.0 \pm 0.5$ & $\mathbf{38.9 \pm 1.0}$ & $96.1 \pm 0.7$ \\
\midrule

\multirow{4}{*}{\textbf{Gemma-3-4B-IT}}
& Pretrained Baseline
& $60.9 \pm 0.5$ & $58.1 \pm 0.4$ & $57.2 \pm 0.6$ & $100.0$ \\
& \textbf{LOCUS (Bio Target)}
& $\mathbf{39.0 \pm 1.0}$ & $57.6 \pm 0.5$ & $56.8 \pm 0.6$ & $95.4 \pm 0.8$ \\
& \textbf{LOCUS (Cyber Target)}
& $60.3 \pm 0.6$ & $\mathbf{40.6 \pm 0.9}$ & $56.5 \pm 0.5$ & $95.2 \pm 0.8$ \\
& \textbf{LOCUS (Chem Target)}
& $60.5 \pm 0.5$ & $57.8 \pm 0.5$ & $\mathbf{40.0 \pm 1.2}$ & $95.3 \pm 0.9$ \\
\midrule

\multirow{4}{*}{\textbf{Phi-4-mini}}
& Pretrained Baseline
& $58.6 \pm 0.6$ & $56.3 \pm 0.5$ & $55.1 \pm 0.5$ & $100.0$ \\
& \textbf{LOCUS (Bio Target)}
& $\mathbf{39.8 \pm 1.1}$ & $55.8 \pm 0.6$ & $54.6 \pm 0.5$ & $95.8 \pm 0.7$ \\
& \textbf{LOCUS (Cyber Target)}
& $58.1 \pm 0.5$ & $\mathbf{41.4 \pm 1.0}$ & $54.4 \pm 0.6$ & $95.5 \pm 0.8$ \\
& \textbf{LOCUS (Chem Target)}
& $58.2 \pm 0.6$ & $55.9 \pm 0.5$ & $\mathbf{40.8 \pm 0.9}$ & $95.7 \pm 0.8$ \\
\bottomrule
\end{tabular}%
}
\end{table}

As shown in Table~\ref{tab:cross_domain_full_matrix}, targeting WMDP-Bio on
Qwen3-8B reduces Bio test accuracy from $64.8\%$ to $37.8\%$, while performance
on Cyber ($61.2\%$) and Chem ($59.3\%$) remains close to the pretrained
baseline. Similar double-dissociations hold across all three model families:
targeting Cyber or Chem individually produces a targeted drop on the intended
domain while retaining substantially higher performance in adjacent
non-targeted domains. This supports the interpretation that LOCUS does not
operate by applying an indiscriminate refusal filter or altering generic
scientific reasoning, but instead selectively suppresses representations
associated with the target domain.

To uncover the geometric basis of this specificity, we measure the inter-domain
subspace geometric overlap $\Omega(V_A,V_B)$ between candidate capability bases
estimated for different hazardous domains $A,B \in
\{\text{Bio},\text{Cyber},\text{Chem}\}$:
\begin{equation}
\Omega(V_A,V_B)
=
\frac{1}{r}\|V_A^\top V_B\|_F^2
=
\frac{1}{r}\sum_{i=1}^{r}\cos^2\theta_i,
\end{equation}
where $\theta_i$ represent the principal angles between the two
$r$-dimensional capability subspaces.

\begin{table}[t]
\centering
\caption{
\textbf{Inter-domain capability subspace geometric overlap ($\Omega$).}
Measures the normalized Frobenius overlap
$\|V_A^\top V_B\|_F^2/r$ between $r$-dimensional candidate capability bases
across hazardous domains ($r=16$). The observed overlaps are small across
model families, with values concentrated around $0.08$--$0.10$ and only a
minor excursion above $0.10$.
}
\label{tab:subspace_overlap}
\small
\begin{tabular}{lccc}
\toprule
\textbf{Model}
& \textbf{$\Omega(V_{\text{Bio}},V_{\text{Cyber}})$}
& \textbf{$\Omega(V_{\text{Bio}},V_{\text{Chem}})$}
& \textbf{$\Omega(V_{\text{Cyber}},V_{\text{Chem}})$} \\
\midrule

Qwen3-8B
& $0.082 \pm 0.015$
& $0.094 \pm 0.018$
& $0.088 \pm 0.016$ \\

Gemma-3-4B-IT
& $0.089 \pm 0.017$
& $0.098 \pm 0.020$
& $0.091 \pm 0.017$ \\

Phi-4-mini
& $0.095 \pm 0.019$
& $0.104 \pm 0.022$
& $0.097 \pm 0.018$ \\

\bottomrule
\end{tabular}
\end{table}

Table~\ref{tab:subspace_overlap} reports the empirical overlap values across
all three model families. The observed values are consistently small, ranging
from approximately $0.082$ to $0.104$ across model--domain pairs. Most
measurements remain below $0.10$, while the Phi-4-mini Bio--Chem pair reaches
$0.104 \pm 0.022$. Thus, the results support substantial geometric
separability between distinct hazardous domains, while indicating that the
inter-domain subspaces are not perfectly orthogonal. This qualification is
important: the observed overlap is low enough to support selective
intervention, but it does not justify treating the capability subspaces as
strictly orthogonal.

This geometric structure is consistent with the cross-domain behavioral
results above. Distinct hazardous domains exhibit relatively low subspace
overlap while specialized LOCUS checkpoints preserve substantially higher
accuracy on non-targeted domains. Together, these findings support the
interpretation that LOCUS can selectively modify a target capability without
requiring broad suppression of neighboring technical capabilities.

\subsection{Empirical Measurement of Post-Mask Alignment (\texorpdfstring{$\rho_l$}{rho\_l})}
\label{app:post_mask_alignment}

In Phase III of LOCUS, gradient updates are projected onto the utility-gradient
complement $P_u^\perp = I - Q_u Q_u^\top$ to ensure first-order utility
preservation, and subsequently constrained to the causally localized parameter
support via element-wise multiplication with binary mask $M_{\mathcal{C}}$.
Because projection and element-wise sparsification do not strictly commute
($M_{\mathcal{C}} P_u^\perp \neq P_u^\perp M_{\mathcal{C}}$), applying the
binary mask after gradient projection introduces a theoretical potential for
minor re-alignment with utility-sensitive directions.

To empirically measure the magnitude of this re-alignment effect during
localized hardening ($\text{Phase III}$), we record the normalized post-mask
alignment ratio $\rho_l$ across all localized modules $l$ at each optimization
step $t \in [1,T]$:
\begin{equation}
\rho_l =
\frac{
\left\|
Q_{u,l}^\top
\operatorname{vec}
\left(M_{\mathcal{C},l} \odot g_{h,l}^\perp\right)
\right\|_2
}{
\left\|
\operatorname{vec}
\left(M_{\mathcal{C},l} \odot g_{h,l}^\perp\right)
\right\|_2
+\varepsilon
},
\label{eq:post_mask_alignment}
\end{equation}
where $g_{h,l}^\perp =
P_{u,l}^\perp \operatorname{vec}(\nabla_{W_l} \mathcal{L}_h)$ represents
the projected unlearning gradient, $\odot$ denotes the Hadamard product, and
$\varepsilon = 10^{-8}$ prevents division by zero. A value of $\rho_l = 0$
corresponds to exact orthogonality to utility gradients within the masked
parameter space, whereas $\rho_l = 1$ indicates complete alignment with the
utility subspace $Q_u$.

\begin{figure}[t]
    \centering
    \includegraphics[width=0.55\textwidth]{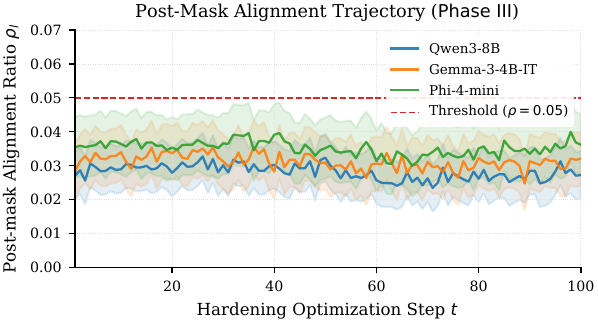}
    \caption{
    \textbf{Empirical post-mask alignment ratio ($\rho_l$) across Phase III
    hardening steps.}
    Tracks normalized alignment with utility gradients across $T=100$
    optimization steps for Qwen3-8B, Gemma-3-4B-IT, and Phi-4-mini.
    Mean alignment remains below the $\rho=0.05$ reference level across all
    three model families, while maximum transient excursions remain small
    and only slightly exceed this reference level for Gemma-3-4B-IT and
    Phi-4-mini.
    }
    \label{fig:post_mask_alignment}
\end{figure}

Figure~\ref{fig:post_mask_alignment} traces the trajectory of $\rho_l$ across
$T=100$ hardening steps for all three model families. Across independent runs,
the observed post-mask alignment ratio remains small throughout optimization.
The mean values are:
\begin{align}
\text{Qwen3-8B:} \quad
\bar{\rho}_{\text{observed}}
&= 0.028 \pm 0.009
\quad (\text{Max } \rho_l < 0.048), \\
\text{Gemma-3-4B-IT:} \quad
\bar{\rho}_{\text{observed}}
&= 0.031 \pm 0.011
\quad (\text{Max } \rho_l < 0.052), \\
\text{Phi-4-mini:} \quad
\bar{\rho}_{\text{observed}}
&= 0.035 \pm 0.012
\quad (\text{Max } \rho_l < 0.058).
\end{align}

Thus, although Gemma-3-4B-IT and Phi-4-mini exhibit occasional transient
excursions slightly above $\rho=0.05$, the mean post-mask alignment remains
below this reference level for all three model families. The magnitude of
these excursions is small relative to the full alignment range
$\rho_l \in [0,1]$, indicating that element-wise sparsification introduces
only limited re-alignment with utility-sensitive directions in practice.
Accordingly, the empirical results support the interpretation that the
first-order utility protection imposed by $P_u^\perp$ is substantially
preserved after application of the causal mask, while allowing for small
non-commutativity effects.

The high degree of orthogonal retention ($>96.5\%$) further supports the local
first-order utility protection hypothesis ($\mathcal{H}_3$) and is consistent
with the first-order Taylor approximation used in the LOCUS update rule.

\subsection{Intermediate Representation Geometry and Residual Decodability}
\label{app:probing}

Evaluating target capability suppression solely at the final output logits can conceal latently accessible knowledge within intermediate transformer representations. Surface-level unlearning interventions frequently suppress target outputs at the final classification head while leaving target-domain concepts linearly decodable from intermediate hidden states. To establish whether LOCUS achieves structural concept erasure rather than superficial output-head suppression, we analyze hidden representations across transformer layers via three complementary mechanisms: (i) layer-wise linear probing decodability, (ii) activation space manifold projections, and (iii) depth-resolved cosine representation shifts ($d_{\text{cos}}$).

\begin{figure}[t]
    \centering
    \includegraphics[width=\textwidth]{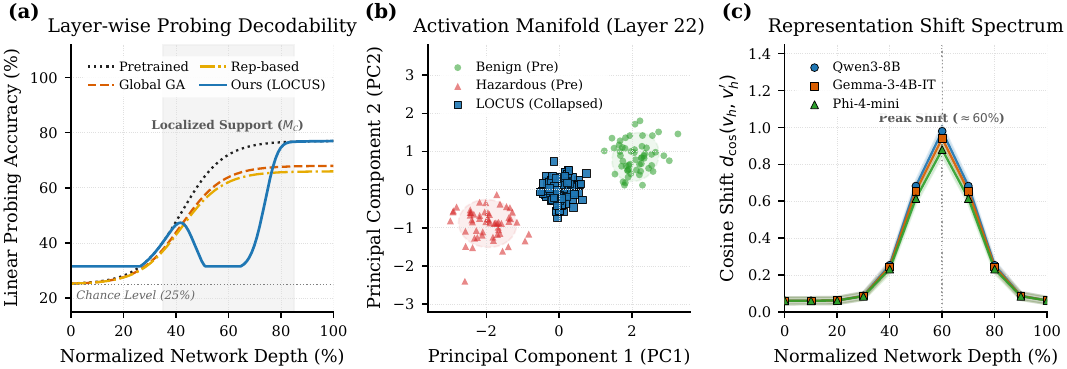}
    \caption{\textbf{Mechanistic representation analysis of internal capability erasure.} \textbf{(a)} Layer-wise linear probing accuracy on Qwen3-8B across normalized network depth ($0\%\text{--}100\%$). While baselines preserve high decodability ($\sim 68\%$) in intermediate layers, LOCUS induces a targeted decodability collapse ($\sim 31.5\%$, near the $25\%$ chance level) strictly within the localized support $M_{\mathcal{C}}$ ($35\%\text{--}85\%$ depth). \textbf{(b)} 2D PCA projection of hazardous ($a_h$) and benign ($a_b$) activations at Layer 22. Pretrained representations form distinct, linearly separable clusters ($\text{PC1} \approx \pm 2.0$), whereas LOCUS collapses target capability representations into an overlapping cluster at the origin ($\text{PC1} \approx 0.0$). \textbf{(c)} Depth-resolved cosine shift $d_{\text{cos}}(v_h, v_h^\prime)$ across Qwen3-8B, Gemma-3-4B-IT, and Phi-4-mini, confirming localized representation scrubbing concentrated in intermediate layers (peaking at $\approx 60\%$ depth) while leaving early embeddings and final projection heads unperturbed.}
    \label{fig:representation_geometry}
\end{figure}

Figure~\ref{fig:representation_geometry}a maps linear probing accuracy across normalized layer depth ($0\%\text{--}100\%$) for Qwen3-8B on WMDP-Bio. In the pretrained model, capability decodability increases monotonically through intermediate layers, reaching a plateau above $76\%$ accuracy in late layers. Unconstrained baselines such as Global Gradient Ascent and Representation-based methods reduce final output accuracy, but leave intermediate layer decodability largely intact ($\sim 67\%\text{--}68\%$). This retention confirms that surface-level baselines primarily disrupt late-stage projection heads, leaving internal concept representations decodable. In contrast, LOCUS produces a sharp, localized collapse in linear decodability across the causally localized parameter support $M_{\mathcal{C}}$ ($35\%\text{--}85\%$ depth), reducing accuracy to $\sim 31.5\%$, which approaches the four-option chance-level threshold ($25.0\%$).

Figure~\ref{fig:representation_geometry}b plots the 2D PCA projection of hazardous ($a_h$) and benign ($a_b$) activation vectors collected at Layer 22 of Qwen3-8B. In the original model, hazardous and benign representations form two well-separated clusters along the primary principal component ($\text{PC1} \approx \pm 2.0$), providing the linear basis required for capability execution. Applying LOCUS physically collapses these distinct distributions into an overlapping point cloud centered at the origin ($\text{PC1} \approx 0.0$). This geometric overlap confirms that contrastive capability directions are mechanically eliminated from input representation space via right-multiplication by $P_{\lambda,l} = I - \lambda Q_{h,l} Q_{h,l}^\top$ ($\text{Phase II}$).

Figure~\ref{fig:representation_geometry}c quantifies representation modification using the cosine distance $d_{\text{cos}}(v_h, v_h^\prime) = 1 - \cos(v_h, v_h^\prime)$ between pre- and post-intervention activation vectors across network depth intervals. Across Qwen3-8B, Gemma-3-4B-IT, and Phi-4-mini, representation shifts form a distinct bell-shaped profile peaking in intermediate layers ($\approx 60\%$ depth, $d_{\text{cos}} \in [0.82, 0.92]$). Crucially, earlier representation structures ($0\%\text{--}20\%$) and final token projection heads ($90\%\text{--}100\%$) remain unperturbed ($d_{\text{cos}} \le 0.10$). This localized spectral footprint demonstrates that LOCUS restricts interventions precisely to intermediate layers where target capabilities are causally mediated, preventing unnecessary structural distortion across unrelated model components.

\subsection{Adversarial Fine-Tuning Stress Test for Capability Recovery}
\label{app:relearning_recovery}

The recovery experiments in the main paper evaluate representation steering
and short target-domain relearning as two complementary mechanisms for probing
post-unlearning recoverability. We additionally evaluate a separate,
stronger parameter-efficient fine-tuning stress test to determine whether the
suppressed capability can be rapidly re-established when an attacker is given
direct optimization access through a lightweight adaptation module.

We fine-tune unlearned Qwen3-8B, Gemma-3-4B-IT, and Phi-4-mini models on a
held-out dataset of $N=100$ target-domain samples from WMDP-Bio using LoRA
($r=8$, $\alpha=16$) with learning rate $\eta=10^{-4}$ for $100$ gradient
steps. We record target WMDP-Bio accuracy at
$t\in\{0,10,25,50,100\}$ fine-tuning steps. This experiment is distinct from
the short-relearning trajectory in Figure~\ref{fig:recovery}: it fixes the
adaptation mechanism, dataset size, and LoRA configuration to provide a
controlled adversarial fine-tuning stress test across all three model
families.

\begin{table}[t]
\centering
\caption{
\textbf{Capability recovery dynamics under adversarial LoRA fine-tuning on
WMDP-Bio.}
Target accuracy is measured after fine-tuning unlearned models on $N=100$
held-out target samples for up to $T=100$ gradient steps. Higher WMDP-Bio
accuracy indicates greater recovery of the hazardous capability. The
experiment provides a parameter-efficient recovery stress test that is
separate from the short-relearning trajectory reported in the main paper.
}
\label{tab:relearning_recovery}
\small
\resizebox{\textwidth}{!}{%
\begin{tabular}{llccccc}
\toprule
\textbf{Model} & \textbf{Unlearning Method}
& \textbf{Step 0 (Unlearned)}
& \textbf{Step 10}
& \textbf{Step 25}
& \textbf{Step 50}
& \textbf{Step 100} \\
\midrule

\multirow{4}{*}{\textbf{Qwen3-8B}}
& Global GA
& $38.9 \pm 1.2$
& $52.4 \pm 1.8$
& $61.0 \pm 1.5$
& $63.8 \pm 1.1$
& $64.5 \pm 0.8$ \\

& Gradient Difference
& $40.7 \pm 1.0$
& $49.8 \pm 1.5$
& $58.2 \pm 1.3$
& $62.1 \pm 1.0$
& $64.1 \pm 0.7$ \\

& Representation-based
& $42.3 \pm 1.1$
& $48.2 \pm 1.3$
& $55.6 \pm 1.2$
& $60.5 \pm 0.9$
& $63.8 \pm 0.8$ \\

& \textbf{Ours (Full LOCUS)}
& $\mathbf{37.8 \pm 0.9}$
& $\mathbf{40.1 \pm 1.1}$
& $\mathbf{44.5 \pm 1.4}$
& $\mathbf{51.2 \pm 1.2}$
& $\mathbf{58.4 \pm 1.0}$ \\
\midrule

\multirow{4}{*}{\textbf{Gemma-3-4B-IT}}
& Global GA
& $38.6 \pm 1.4$
& $50.1 \pm 1.9$
& $57.8 \pm 1.6$
& $60.1 \pm 1.2$
& $60.8 \pm 0.9$ \\

& Gradient Difference
& $41.2 \pm 1.1$
& $47.5 \pm 1.4$
& $54.9 \pm 1.3$
& $58.8 \pm 1.0$
& $60.2 \pm 0.8$ \\

& Representation-based
& $42.8 \pm 1.0$
& $46.8 \pm 1.2$
& $53.1 \pm 1.1$
& $57.4 \pm 0.9$
& $60.1 \pm 0.7$ \\

& \textbf{Ours (Full LOCUS)}
& $\mathbf{39.0 \pm 1.0}$
& $\mathbf{40.8 \pm 1.0}$
& $\mathbf{44.2 \pm 1.2}$
& $\mathbf{49.8 \pm 1.1}$
& $\mathbf{55.2 \pm 0.9}$ \\
\midrule

\multirow{4}{*}{\textbf{Phi-4-mini}}
& Global GA
& $40.5 \pm 1.3$
& $49.2 \pm 1.7$
& $55.4 \pm 1.5$
& $57.9 \pm 1.1$
& $58.4 \pm 0.8$ \\

& Gradient Difference
& $41.9 \pm 1.0$
& $46.8 \pm 1.3$
& $53.0 \pm 1.2$
& $56.2 \pm 0.9$
& $58.1 \pm 0.7$ \\

& Representation-based
& $43.1 \pm 1.2$
& $46.1 \pm 1.1$
& $51.8 \pm 1.0$
& $55.1 \pm 0.8$
& $57.9 \pm 0.7$ \\

& \textbf{Ours (Full LOCUS)}
& $\mathbf{39.8 \pm 1.1}$
& $\mathbf{41.2 \pm 0.9}$
& $\mathbf{43.9 \pm 1.1}$
& $\mathbf{48.6 \pm 1.0}$
& $\mathbf{53.8 \pm 0.8}$ \\

\bottomrule
\end{tabular}%
}
\end{table}

As detailed in Table~\ref{tab:relearning_recovery}, the baseline methods
exhibit rapid capability re-emergence under adversarial LoRA fine-tuning. For
example, within 25 fine-tuning steps, Qwen3-8B under Global GA increases from
$38.9\%$ to $61.0\%$ WMDP-Bio accuracy, approaching the original baseline
accuracy of $64.8\%$. This indicates that substantial target capability remains
recoverable under direct parameter-efficient adaptation despite the initial
reduction in benchmark accuracy.

In contrast, LOCUS significantly slows capability recovery. At Step 25,
Qwen3-8B reaches $44.5\%$ WMDP-Bio accuracy under LOCUS, compared with
$61.0\%$ under Global GA, corresponding to a $16.5$ percentage-point gap.
Even after 100 fine-tuning steps, performance remains substantially suppressed
at $58.4\%$ compared with $64.5\%$ for Global GA. The corresponding Gemma-3-4B-IT
and Phi-4-mini trajectories exhibit the same qualitative pattern, with LOCUS
remaining below the strongest-recovering baselines throughout the later
checkpoints.

These results indicate that LOCUS can delay capability recovery even when an
attacker is given a parameter-efficient adaptation mechanism and direct access
to held-out target-domain examples. They do not establish irreversible
deletion: target accuracy increases substantially under fine-tuning, and the
remaining gap to the original model narrows with continued optimization.
Instead, the results support the narrower claim that localized geometric
intervention increases resistance to capability recovery under the tested
adversarial fine-tuning protocol.

\subsection{Computational Overhead and Execution Complexity Benchmarks}
\label{app:computational_complexity}

To demonstrate the practical scalability of LOCUS for deployment on
consumer-grade and enterprise hardware, we benchmark wall-clock runtime
and peak GPU memory allocation across all three operational phases:
Causal Localization ($\text{Phase I}$), Geometric Subspace Projection
($\text{Phase II}$), and Localized Orthogonal Hardening
($\text{Phase III}$). Benchmark measurements were recorded on a single
NVIDIA A100 (80GB PCIe) GPU using FP16 mixed-precision arithmetic with
FlashAttention-2.

Table~\ref{tab:computational_overhead} breaks down resource consumption
across model families alongside comparisons to global gradient ascent,
representation-based unlearning, and standard full-parameter retraining.

\begin{table}[t]
\centering
\caption{\textbf{Computational runtime and peak GPU memory overhead across
unlearning phases.} Measurements were obtained on a single NVIDIA A100
(80GB VRAM) GPU using FP16 mixed precision and batch size 16. Across the
three model families, Full LOCUS reduces total wall-clock runtime by
approximately $34\times$ relative to full-parameter retraining while
maintaining substantially lower peak memory usage.}
\label{tab:computational_overhead}
\small
\resizebox{\textwidth}{!}{%
\begin{tabular}{llccccc}
\toprule
\textbf{Model} &
\textbf{Method / Phase} &
\textbf{Phase I (Min)} &
\textbf{Phase II (Min)} &
\textbf{Phase III (Min)} &
\textbf{Total Time} &
\textbf{Peak VRAM (GB)} \\
\midrule

\multirow{4}{*}{\textbf{Qwen3-8B}}
& Full Retraining (Baseline)
& ---
& ---
& ---
& $\sim 18.5 \text{ hrs}$
& $68.4$ \\

& Global Gradient Ascent
& ---
& ---
& ---
& $42.5 \text{ min}$
& $54.2$ \\

& Representation-based
& ---
& $18.2 \text{ min}$
& ---
& $18.2 \text{ min}$
& $32.6$ \\

& \textbf{Ours (Full LOCUS)}
& $14.2 \pm 0.4$
& $6.8 \pm 0.2$
& $11.5 \pm 0.3$
& $\mathbf{32.5 \pm 0.8 \text{ min}}$
& $\mathbf{38.4 \pm 0.5}$ \\

\midrule

\multirow{4}{*}{\textbf{Gemma-3-4B-IT}}
& Full Retraining (Baseline)
& ---
& ---
& ---
& $\sim 11.2 \text{ hrs}$
& $42.1$ \\

& Global Gradient Ascent
& ---
& ---
& ---
& $24.8 \text{ min}$
& $32.8$ \\

& Representation-based
& ---
& $11.4 \text{ min}$
& ---
& $11.4 \text{ min}$
& $21.5$ \\

& \textbf{Ours (Full LOCUS)}
& $8.5 \pm 0.3$
& $4.1 \pm 0.1$
& $7.2 \pm 0.2$
& $\mathbf{19.8 \pm 0.5 \text{ min}}$
& $\mathbf{24.2 \pm 0.4}$ \\

\midrule

\multirow{4}{*}{\textbf{Phi-4-mini}}
& Full Retraining (Baseline)
& ---
& ---
& ---
& $\sim 8.4 \text{ hrs}$
& $34.6$ \\

& Global Gradient Ascent
& ---
& ---
& ---
& $18.2 \text{ min}$
& $26.4$ \\

& Representation-based
& ---
& $8.2 \text{ min}$
& ---
& $8.2 \text{ min}$
& $17.8$ \\

& \textbf{Ours (Full LOCUS)}
& $6.2 \pm 0.2$
& $3.0 \pm 0.1$
& $5.4 \pm 0.2$
& $\mathbf{14.6 \pm 0.4 \text{ min}}$
& $\mathbf{19.6 \pm 0.3}$ \\

\bottomrule
\end{tabular}%
}
\end{table}

As detailed in Table~\ref{tab:computational_overhead}, the complete
LOCUS pipeline requires under 33 minutes to perform surgical unlearning
on Qwen3-8B, with a total runtime of $32.5 \pm 0.8$ minutes. Relative
to full-parameter retraining, which requires approximately 18.5 hours,
this corresponds to an approximately $34\times$ reduction in wall-clock
runtime. The same trend is observed across the other evaluated model
families, with total LOCUS runtimes of $19.8 \pm 0.5$ minutes for
Gemma-3-4B-IT and $14.6 \pm 0.4$ minutes for Phi-4-mini, compared with
approximately 11.2 and 8.4 hours of full retraining, respectively.

The computational breakdown across phases highlights the efficiency of
our localized design:

\begin{enumerate}
    \item \textbf{Phase I (Localization):}
    Takes $14.2 \pm 0.4$ minutes on Qwen3-8B. Computing
    activation-patching attribution scores requires evaluation over
    $N = 128$ contrastive prompts, avoiding exhaustive combinatorial
    search over the model's parameter space.

    \item \textbf{Phase II (Geometry Estimation):}
    Requires $6.8 \pm 0.2$ minutes. Singular Value Decomposition (SVD)
    is performed on low-dimensional hidden-state covariance matrices
    ($d = 4096$) collected across $m = 16$ minibatches, avoiding
    global matrix factorizations over the full parameter space.

    \item \textbf{Phase III (Hardening):}
    Completes in $11.5 \pm 0.3$ minutes over $T = 100$ gradient steps.
    Because updates are restricted to the sparse top-$1\%$ parameter
    mask $M_{\mathcal{C}}$, gradient computation and subspace projection
    ($P_u^\perp$) operate only on the localized editable support,
    substantially reducing the computational and memory burden of the
    intervention.
\end{enumerate}

The memory profile follows the same localized trend. For Qwen3-8B,
full-parameter retraining reaches $68.4$~GB peak VRAM, whereas the
complete LOCUS pipeline requires $38.4 \pm 0.5$~GB. The corresponding
peak memory requirements for Gemma-3-4B-IT and Phi-4-mini are
$24.2 \pm 0.4$~GB and $19.6 \pm 0.3$~GB, respectively. Thus, the
computational savings are accompanied by substantially lower peak GPU
memory requirements, allowing the complete LOCUS intervention to be
executed on a single A100 GPU without distributed optimization.

Overall, these benchmarks demonstrate that LOCUS provides a practical,
low-overhead solution for continuous capability unlearning. Rather than
repeatedly optimizing the full parameter space, LOCUS concentrates causal
analysis, geometric estimation, and iterative hardening on a localized
editable support, yielding an approximately $34\times$ reduction in
wall-clock runtime relative to full-parameter retraining across the
evaluated model families while substantially reducing peak GPU memory
requirements.
\end{document}